\documentclass{article}

\usepackage[nonatbib, preprint]{neurips_2020}

\usepackage[utf8]{inputenc} 
\usepackage[T1]{fontenc}    
\usepackage{hyperref}       
\usepackage{url}            
\usepackage{booktabs}       
\usepackage{amsfonts}       
\usepackage{nicefrac}       
\usepackage{microtype}      
\usepackage{graphicx}
\usepackage{subfigure}
\usepackage{algorithm}

\usepackage{url}
\usepackage{breakurl}
\usepackage{algorithmic}

\usepackage{amsmath}
\usepackage{amssymb}
\usepackage{color}

\title{On the Promise of the Stochastic Generalized Gauss-Newton Method for Training DNNs}

\author{Matilde Gargiani$^1$, Andrea Zanelli$^{3}$, Moritz Diehl$^{3,4}$, Frank Hutter$^{2,5}$\\ 
$^1$Automatic Control Laboratory, ETH Zurich\\
\texttt{gmatilde@control.ee.ethz.ch} \\
$^2$Department of Computer Science, University of Freiburg\\
\texttt{fh@cs.uni-freiburg.de} \\
$^3$Department of Microsystems Engineering (IMTEK), University of Freiburg\\
\texttt{\{andrea.zanelli, moritz.diehl\}@imtek.uni-freiburg.de} \\
$^4$Department of Mathematics, University of Freiburg\\
$^5$Bosch Center for Artificial Intelligence
}

\author{%
  Matilde Gargiani\thanks{This work was done while the author was working at the University of Freiburg.} \\
  Automatic Control Laboratory\\
  ETH Zurich\\
  \texttt{gmatilde@control.ee.ethz.ch} \\
   \And
   Andrea Zanelli \\
   Department of Microsystems Engineering\\
   University of Freiburg\\
   \texttt{andrea.zanelli@imtek.uni-freiburg.de} \\
   \AND
   $\!\!\!\!\!\!\!\!\!\!\!\!\!\!\!\!\!\!\!$Moritz Diehl \\
   $\!\!\!\!\!\!\!\!\!\!\!\!\!\!\!\!$Department of Microsystems Engineering\\
   $\!\!\!\!\!\!\!\!\!\!\!\!\!\!\!\!$and Department of Mathematics\\
   $\!\!\!\!\!\!\!\!\!\!\!\!\!\!\!\!$University of Freiburg \\
   $\!\!\!\!\!\!\!\!\!\!\!\!\!\!\!\!$\texttt{moritz.diehl@imtek.uni-freiburg.de} \\
   \And
   $\!\!\!\!\!\!\!\!\!\!\!\!\!\!$Frank Hutter \\
   $\!\!\!\!\!\!\!\!\!\!\!\!\!\!\!$Department of Computer Science\\
   $\!\!\!\!\!\!\!\!\!\!\!\!\!\!\!$University of Freiburg \\
   $\!\!\!\!\!\!\!\!\!\!\!\!\!\!\!$and Bosch Center for Artificial Intelligence\\
   $\!\!\!\!\!\!\!\!\!\!\!\!\!\!\!$\texttt{fh@cs.uni-freiburg.de}\\
}

\begin{document}
\maketitle

\begin{abstract}
Following early work on Hessian-free methods for deep learning, we study a stochastic generalized Gauss-Newton method (SGN) for training deep neural networks. SGN is a second-order optimization method, with efficient iterations, that we demonstrate to often require substantially fewer iterations than standard SGD to converge. As the name suggests, SGN uses a Gauss-Newton approximation for the Hessian matrix, and, in order to efficiently compute an approximate search direction, relies on the conjugate gradient method combined with forward and reverse automatic differentiation. Despite the success of SGD and its first-order variants in deep learning applications, and despite Hessian-free methods based on the Gauss-Newton Hessian approximation having been already theoretically proposed as practical methods for training neural networks, we believe that SGN has a lot of undiscovered and yet not fully displayed potential in big mini-batch scenarios. For this setting, we demonstrate that SGN does not only substantially improve over SGD in terms of the number of iterations, but also in terms of runtime. This is made possible by an efficient, easy-to-use and flexible implementation of SGN we propose in the Theano deep learning platform, which, unlike Tensorflow and Pytorch, supports forward automatic differentiation. This enables researchers to further study and improve this promising optimization technique and hopefully reconsider stochastic second-order methods as competitive optimization techniques for training DNNs; we also hope that the promise of SGN may lead to forward automatic differentiation being added to Tensorflow or Pytorch. Our results also show that in big mini-batch scenarios SGN is more robust than SGD with respect to its hyperparameters (we never had to tune its step-size for our benchmarks!), which eases the expensive process of hyperparameter tuning that is instead crucial for the performance of first-order methods. 
\end{abstract}

\section{Introduction}
The training of a deep neural network requires the solution of a large scale nonconvex unconstrained optimization problem, and it is typically addressed with first-order methods: the workhorse optimization algorithms used for training are indeed stochastic gradient descent~\cite{sgd_bottou} and its variants~\cite{polyak:1964},~\cite{nesterov:1983}. 
These methods are characterized by cheap iterations, but their iterates generally progress very slowly, especially in the presence of pathological curvature regions. In addition, they require time and resource demanding tuning procedures for adjusting the learning rate value in the course of the optimization~\cite{Goodfellow-et-al-2016}. Second-order methods, which include curvature information in their update rules, would mitigate the effect of poor-conditioning, but at the price of more expensive iterations.  In practice, these methods automatically adjust their search direction and step-size according to the quadratic local approximation of the objective. 
 
In this work, we consider unconstrained optimization problems of the form
\begin{equation}
\label{eq: gen_problem}
{\arg}\min_{\boldsymbol{\theta} \in \mathbb{R}^{d}}\,\,L \left(\boldsymbol{\theta} \right),
\end{equation}
where $d\gg 0$ and $L:\mathbb{R}^{d}\rightarrow \mathbb{R}$ is a twice differentiable function that is nonconvex in general. We now briefly review some of the second-order methods that have been proposed in the literature to address the solution of Problem~\eqref{eq: gen_problem}. Typically, at each iteration, these methods require the solution of a non-trivial large scale linear system
\begin{equation}
\label{eq: lin_system}
B_k \Delta \boldsymbol{\theta}_k = -\mathbf{g}_k,
\end{equation} 
where $\mathbf{g}_k \in \mathbb{R}^{d}$ is the local gradient, and $B_k \in \mathbb{R}^{d\times d}$ a matrix of (estimated) local curvature information.
In the scenario of deep learning, computing explicitly, storing and factorizing this matrix is generally intractable or, at least, undesirable from a computational viewpoint. In addition, due to the large number of parameters in the system, it is indeed necessary to reduce to at most matrix-vector multiplications the computational complexity of the optimization algorithms.   
Schraudolph~\cite{schraudolph02} has been among the first in the deep learning community to propose a general method for iteratively approximating various second-order steps (i.e. Newton, Gauss-Newton, Levenberg-Marquardt and natural gradient). In particular, in~\cite{schraudolph02} he proposes to solve the linear system~\eqref{eq: lin_system} by applying gradient-descent to the following minimization problem
\begin{equation}
\label{eq: quadr_problem}
{\arg}\min_{\Delta\boldsymbol{\theta}_k}\,\, \mathbf{g}_k^\top \Delta \boldsymbol{\theta}_k+\frac{1}{2}\, \Delta \boldsymbol{\theta}_k^\top B_k \Delta \boldsymbol{\theta}_k,
\end{equation} 
and he suggests to use it in conjunction with automatic differentiation (AD) techniques for computing curvature matrix-vector products, in order to design rapid, iterative and optionally stochastic variants of second-order methods. 
However, the use of a gradient method to solve~\eqref{eq: quadr_problem} is an approach that is likely to suffer from slow convergence since~\eqref{eq: quadr_problem} exhibits the same pathological curvature of the objective function that it locally approximates.
The work of Martens on Hessian-free methods~\cite{icml2010_094} is also relevant to our analysis; it is directly built on previous work of Pearlmutter on AD~\cite{pearlmutter_ad} and fast exact multiplication by the Hessian~\cite{Pearlmutter:93}, and the work of Schraudolph on fast second-order methods~\cite{schraudolph02}. Martens~\cite{Martens2014NewIA} proposes a second-order optimization method based on the so-called Hessian-free approach (see~\cite{NoceWrig06} for more details): in particular, he proposes the use of the $\mathcal{R}\left\{ \cdot \right\}$ operator (see~\cite{Pearlmutter:93} for more details) to efficiently compute the exact Hessian-vector product without explicitly computing the Hessian matrix, and then to solve~\eqref{eq: lin_system} with the Conjugate Gradient (CG) method~\cite{NoceWrig06}. However, because of the use of the exact Hessian, the method must rely on a regularization technique to ensure positive definiteness of the Hessian. 
On the contrary, the Gauss-Newton Hessian approximation is  positive semidefinite by construction~\cite{NoceWrig06}, and therefore it can theoretically be used directly with CG for any positive value of the regularization parameter. The advantages of using a Gauss-Newton type approximation for the Hessian-matrix and its relation to the Fisher matrix are underlined by Martens in his work on the natural gradient method~\cite{Martens2014NewIA}.
The K-FAC algorithm~\cite{DBLP:journals/corr/MartensG15} is also a relevant large-scale stochastic second order method for training DNNs. The method is based on an approximation of the Fisher information matrix to capture the curvature information at each iteration. In particular, a block-diagonal structure is imposed to the Fisher matrix such that, together with an additional approximation based on the Kronecker-factorizations, its inverse is efficiently and directly computed without the use of iterative methods that are generally deployed to solve~\eqref{eq: lin_system}. 
\subsection{Contributions}\label{subsec:contributions}
Taking inspiration from previous work on Hessian-free and Gauss-Newton methods for deep learning, in this work we study across different benchmarks the performance of an Hessian-free method called SGN that combines the use of
\begin{itemize}
\item a generalized Gauss-Newton Hessian approximation to capture curvature information, and
\item the conjugate gradient (CG) method together with AD, in both forward and reverse mode, to efficiently obtain an approximate Gauss-Newton direction. 
\end{itemize}
Even if these techniques are widely used in the optimization community, to the best of our knowledge, their potential for training deep neural networks is still not fully discovered. This is partly due to the complexity of implementing SGN and the lack in the main deep learning frameworks, Pytorch~\cite{pytorch_2019} and Tensorflow~\cite{10.5555/3026877.3026899}, of an exhaustive implementation of forward AD, on which Hessian-free operations rely. Our work aims at offering a first ready- and easy-to-use tool to discover the potential of SGN. In particular, our work consists in 
\begin{itemize}
\item a mathematical description of the method,
\item an efficient implementation of SGN in Theano~\cite{theano}, specifically optimized in order to use the solver for training DNNs and with many available features,
\item numerical results on different benchmarks showing that SGN is competitive with state-of-the-art methods for training deep neural networks.
\end{itemize}
Our benchmarks show that in big mini-batch scenarios SGN can lead to faster convergence and better generalization than SGD.  
We hope that these results can help to open the eyes of the deep learning community on the need for more sophisticated optimization methods especially with the advent of modern graphics processing units that allow for massive parallelization and therefore are designed to work in big mini-batch scenarios. 
\section{Method}
\subsection{Problem Formulation}
Given a deep neural network with $d\gg 0$ parameters, $n$ inputs and $m$ outputs, its output $\mathbf{y}$ can be represented as the composition of two functions $\mathbf{y} = \left(o \circ f\right)\left( \mathbf{x}; \boldsymbol{\theta} \right)$, where $f:\mathbb{R}^{d}\rightarrow\mathbb{R}^{m}$ is the function describing the network except for the nonlinearity of the last layer, and $o:\mathbb{R}^{m}\rightarrow\mathbb{R}^{m}$ corresponds to the activation function of the output layer. The vector of parameters is denoted by $\boldsymbol{\theta} \in \mathbb{R}^d$, and $\mathbf{x}\in \mathbb{R}^n$ represents the input vector.\medskip \newline The so-called training process of a neural network aims at finding the best value of the parameters $\boldsymbol{\theta}$ by minimization of a nonconvex loss function $L :\mathbb{R}^{d}\rightarrow \mathbb{R}$ that is typically obtained by considering the log-likelihood of the network's output, here denoted as $\ell: \mathbb{R}^{m}\rightarrow \mathbb{R}$, under a suitable statistical model. In particular, given that the data distribution is unknown, but we only dispose of $N\gg 0$ realizations of it, the large scale nonconvex stochastic optimization problem that we aim to solve is
\begin{equation}\label{eq:prob2}
\begin{aligned}
{\arg}\min_{\boldsymbol{\theta}} \,\, L(\boldsymbol{\theta})\,=\, & {\arg}\min_{\boldsymbol{\theta}}
\,\, \frac{1}{N}\,\mathop{\sum_{i=1}^{N}} \ell (\bar{y}_i\, ; \, \mathbf{y}_i) \, ,
\end{aligned}
\end{equation}
where $\bar{y}_i  \in \mathbb{R}$ is the label of the sample ${\mathbf{x}_i}$. We have used the notation $\mathbf{y}_i = \left(o\circ f_i\right) ( \boldsymbol{\theta})$ to denote $\mathbf{y}_i = \left(o\circ f\right) (\mathbf{x}_i; \boldsymbol{\theta})$.

Typical choices for the activation function of the output layer and the loss function are: softmax function combined with the cross-entropy loss for classification problems, and identity map with mean squared error for regression tasks~\cite{Goodfellow-et-al-2016}. In both cases, $\ell \circ o$  is a convex function.

By applying the chain-rule, we can easily obtain the Jacobian
\begin{equation}\label{eq:jacobian}
\begin{aligned}
& \mathit{J}_{L}(\boldsymbol{\theta})\,= \, \frac{1}{N}\, \mathop{\sum_{i=1}^{N}} \mathit{J}_{\ell\,\circ\,o}(\mathbf{y}_i)\cdot  \mathit{J}_{f_i}(\boldsymbol{\theta}),
\end{aligned} 
\end{equation}

and the Hessian
\begin{equation}\label{eq:hessian}
\begin{aligned}
& H_{L}(\boldsymbol{\theta}) \,= \, \frac{1}{N} \, \mathop{\sum_{i=1}^{N}} J_{f_i}^\top(\boldsymbol{\theta}) \cdot  H_{\ell \,\circ\, o}(\mathbf{y}_i)\cdot  J_{f_i} (\boldsymbol{\theta})+ H_{f_i}(\boldsymbol{\theta})\otimes J_{\ell\,\circ\, o}(\mathbf{y}_i),
\end{aligned} 
\end{equation}
of the objective function of problem~\eqref{eq:prob2}.
\subsection{Generalized Gauss-Newton Hessian Approximation}
\label{GGN}
The Gauss-Newton Hessian is specifically designed for nonlinear least-squares objectives
\begin{equation}
f(x) = \frac{1}{2} \underset{j=1}{\overset{k}\sum}\, r_j^2(x),
\label{eq:nl_leastsq}
\end{equation}
where each $r_j:\mathbb{R}^z\rightarrow \mathbb{R}$ is a smooth function called residual. 
By introducing the residual vector
\begin{equation}
\mathbf{r}(x)= (r_1(x),\dots,r_k(x))^\top,
\end{equation}
we can then write an equivalent formulation of~\eqref{eq:nl_leastsq} as
\begin{equation}
f(x) = \frac{1}{2}||\mathbf{r}(x)||^2.
\end{equation}

The gradient and Hessian matrix can be expressed as follows:
\begin{align}
&\nabla f(x) = J(x)^\top \mathbf{r}(x), \\
\label{eq:nl_hessian}
&\nabla^2 f(x) = J(x)^\top J(x)+\underset{j=1}{\overset{k}{\sum}}r_j(x)\nabla^2r_j(x),
\end{align}
where $J(x)$ is the Jacobian matrix of the residuals.
The Gauss-Newton approximation consists in  
\begin{align}\label{eq:nl_GN}
&\nabla^2 f(x) \approx J(x)^\top J(x),
\end{align}
which is by construction positive semidefinite and it makes it possible to avoid the computation of the residual Hessians in~\eqref{eq:nl_hessian}. 
The Hessian approximation~\eqref{eq:nl_GN} is particularly accurate in the case where the residuals $\mathbf{r}(x)$ are ``small'' relatively to the first term at the linearization point~\cite{NoceWrig06}. 
Function~\eqref{eq:nl_leastsq} can also be intended as composition of the squared 2-norm function with the residual function $r:\mathbb{R}^n\rightarrow\mathbb{R}^m$. If we replace the squared 2-norm function with any other arbitrary convex function,  then we can still apply the same principles of the Gauss-Newton approximation: this extension is known as generalized Gauss-Newton method~\cite{generalized_gn}. 

In the case of deep neural networks, the generalized Gauss-Newton approximation of~\eqref{eq:hessian} is given by
\begin{equation}
\mathbf{H}^{\mathrm{GN}}_{L}(\boldsymbol{\theta})  \,= \, \frac{1}{N} \, \mathop{\sum_{i=1}^{N}} J_{f_i}^\top(\boldsymbol{\theta}) \cdot  H_{\ell \,\circ\, o}(\mathbf{y}_i)\cdot  J_{f_i} (\boldsymbol{\theta})\, .
\label{eq:GN_hessian}
\end{equation}
Assuming $\ell \circ o$ to be a convex function, then $\mathbf{H}^{\mathrm{GN}}_{L}(\boldsymbol{\theta})\succeq 0$. 

In order to make our method more robust and ensure positive definiteness, we make use of a  Levenberg-Marquardt Hessian approximation~\cite{NoceWrig06}, which additionally incorporates a regularization term $\rho I \in \mathbb{R}^{d\times d}$:
\begin{equation}
\mathbf{H}^{\mathrm{GN}}_{L}(\boldsymbol{\theta})  \,= \, \frac{1}{N} \, \mathop{\sum_{i=1}^{N}} J_{f_i}^\top(\boldsymbol{\theta}) \cdot  H_{\ell \,\circ\, o}(\mathbf{y}_i)\cdot  J_{f_i} (\boldsymbol{\theta}) + \rho I,
\label{eq:LMGN_hessian}
\end{equation}
where $\rho$ can be interpreted as a tuning parameter in a trust region framework. 

The search direction is given by the solution of the following linear system
\begin{equation}
\left(\mathbf{H}^{\mathrm{GN}}_{L}(\boldsymbol{\theta})+\rho I\right)\Delta\boldsymbol{\theta} = -J_L^\top(\boldsymbol{\theta}) \,.
\label{eq:linear_system}
\end{equation}

In a large scale scenario, such as the one of deep learning, it is often not feasible to compute, store and factorize directly~\eqref{eq:GN_hessian}. Therefore, in order to solve~\eqref{eq:linear_system}, we opt for the use of CG method, a well known iterative method for solving large scale linear systems with symmetric positive definite matrices~\cite{NoceWrig06}. CG allows not only to avoid the factorization, but also the direct computation and storage of~\eqref{eq:GN_hessian}, since it only requires Hessian-vector products. Even if, in the worst-case scenario, CG requires $d$ iterations to find the solution, it is known that its convergence rate strongly depends on the distribution of Hessian approximation's eigenvalues: if the eigenvalues are clustered, then CG will provide a good approximate solution after a number of steps that is about the same as the number of clusters~\cite{NoceWrig06}. Given that $d\gg0$ for the applications of our interest, it would not be feasible to perform the CG iterations until convergence in the worst-case scenario, nor really useful for the optimization given the stochastic setting. 
However, given the empirical evidence of clustered eigenvalues for deep networks~\cite{dauphin:2014} and since we are only interested in finding an approximate solution of~\eqref{eq:linear_system}, we perform a small number of iterations. See Algorithm~\ref{alg:example} in Section~\ref{sec:algorithms} of the Appendix for a pseudo code description of SGN.  

\subsection{Automatic Differentiation}
Thanks to the structure of the CG iterations (see Algorithm~\ref{alg:CG_alg} in Section~\ref{sec:algorithms} of the Appendix), we only need to evaluate the product of our Hessian approximation with a vector: this can be efficiently done by exploiting AD in its forward and reverse accumulation modes~\cite{moritz}. In particular, forward mode AD provides a very efficient and matrix-free way of computing the Jacobian-vector product $J_{f_i}(\boldsymbol{\theta}) \cdot \mathbf{v}$: 
\begin{equation}
\mathrm{cost}_{\mathrm{FAD}}\left(J_{f_i}(\boldsymbol{\theta}) \cdot \mathbf{v}\right) \,\leq\, 2 \,\mathrm{cost}(f_i)\,,
\end{equation}
and, similarly, the reverse mode can be used to compute the transposed Jacobian-vector product $J_{f_i}^\top (\boldsymbol{\theta}) \cdot \mathbf{u}$
\begin{equation}
\mathrm{cost}_{\mathrm{RAD}} \left(J_{f_i}^\top (\boldsymbol{\theta}) \cdot \mathbf{u}\right) \,\leq\, 3\, \mathrm{cost}(f_i)\,, 
\end{equation}
where $\mathrm{cost}(f_i)$ is the cost of a forward evaluation of $f_i$. 
Finally, the computation of the curvature matrix-vector product can be fully parallelized with respect to the data samples. 
\subsection{Handling Stochasticity}
In order to handle large datasets ($N\gg0$), such as those typically used in deep learning applications, we adopt the same stochastic setting as the other state-of-the-art optimization methods specifically designed and used for training neural networks~\cite{Goodfellow-et-al-2016}. Consequently, in each iteration, both the quantities~\eqref{eq:jacobian} and~\eqref{eq:GN_hessian} are not computed over the entire dataset but estimated over a mini-batch of only $M\ll N$ samples, randomly selected from the training data. Since in this case we are also estimating second-order type of information, the mini-batch size generally needs to be bigger than that normally used for stochastic first-order methods.
\section{Theano Implementation}
We implemented SGN in Theano~\cite{theano} in the form of an easy-to-use Python package with different available features. The choice of Theano, as already discussed in~\ref{subsec:contributions}, is due to the fact that, as the time of writing, it is the only deep learning framework to provide an efficient and fast way to compute Jacobian-vector and vector-Jacobian products via the $\mathcal{L} \left\{\cdot\right\}$ and $\mathcal{R} \left\{\cdot\right\}$ operators. Our code, together with the results of our benchmarks and explicative code examples, is publicly available at \url{https://github.com/gmatilde/SGN}.

\subsection{Code Structure and Available Features}
The class \verb|SGN| with its methods allows for the instantiation and deployment of SGN as optimizer for training a network. The code and available examples at \url{https://github.com/gmatilde/SGN} are self-explanatory. The implementation mimics in its structure the Pytorch implementation of optimizers such as SGD~\cite{sgd_bottou} and Adam~\cite{adam} with the main difference that, to ease the use of the $\mathcal{L} \left\{\cdot\right\}$ and $\mathcal{R} \left\{\cdot\right\}$ operators, the model's parameters are required to be reshaped in a single vector. For this reason, the package also comes together with the classes \verb|mlp| and \verb|vgg| which allow to easily create feedforward and vgg-type of neural networks that can be directly handled by the \verb|SGN| class.

Our implementation of SGN comes together with different features. The code documentation available at \url{https://github.com/gmatilde/SGN} lists and explains them in details. Among those, there is also the possibility of using backtracking line search (see Algorithm 3.1 in~\cite{NoceWrig06}) to adaptively adjust the step lengths based on the so-called sufficient decrease condition, as well as the possibility of activating a trust-region based approach to adaptively adjust the Levenberg-Marquardt parameter $\rho$ (see Algorithm 4.1 in~\cite{NoceWrig06}) based on the ratio of the so-called actual and predicted reductions (see Chapter 4 in~\cite{NoceWrig06}). Alternatively to the adaptive trust-region based approach and inspired by the learning rate decaying schedules generally used for SGD, our SGN implementation also offers the possibility of using different non-adaptive decaying schedules for the Levenberg-Marquardt parameter, such as a linear or cosine decay schedules.    
\section{Experiments}
In this Section, we conduct an extensive empirical analysis of the performance of SGN in big mini-batch scenarios. In particular, we compare its performance with different values of CG iterations to that of SGD with different values of learning rate on different datasets: boston housing from the UCI machine learning repository~\cite{Dua_2019}, samples drawn from a sine wave function corrupted by Gaussian noise, MNIST~\cite{lecun2010mnist}, FashionMNIST~\cite{xiao2017} and CIFAR10~\cite{Krizhevsky09learningmultiple}.

Except for the number of CG iterations, SGN is run with an initial value of $\rho=10^{-4}$ and the default values for the other hyperparameters in all the benchmarks (see the documentation at \url{https://github.com/gmatilde/SGN}).

\subsection{Boston Housing}
As regressing model for this task we use a simple feedforward neural network with 2 hidden layers of 100 units each and sigmoid as activation function. We set the mini-batch size to $101$, which corresponds to one fourth of the training set size (404). As objective function we use the standard mean squared error loss.  
As shown in the left plot of Figure~\ref{fig:boston_sine_time} and in Figures~\ref{fig:boston_1}-\ref{fig:boston_4} in the Appendix, SGD's iterates for different values of the learning rate get stacked in a bad-minimizer, while SGN's iterates quickly converge to a better minimizer with lower values of train and test loss. Regarding SGD, the results with learning rate value of 1 are not shown in the Figures for clarity, as with this learning rate value SGD's iterates quickly diverge leading to high values of the objective function. These benchmarks are run on a Intel(R) Core(TM) i7-7560U CPU @ 2.40GHz. 
\begin{figure}[!htb]
    \centering
    \begin{minipage}{.5\textwidth}
        \centering
        {\includegraphics[width=1\textwidth]{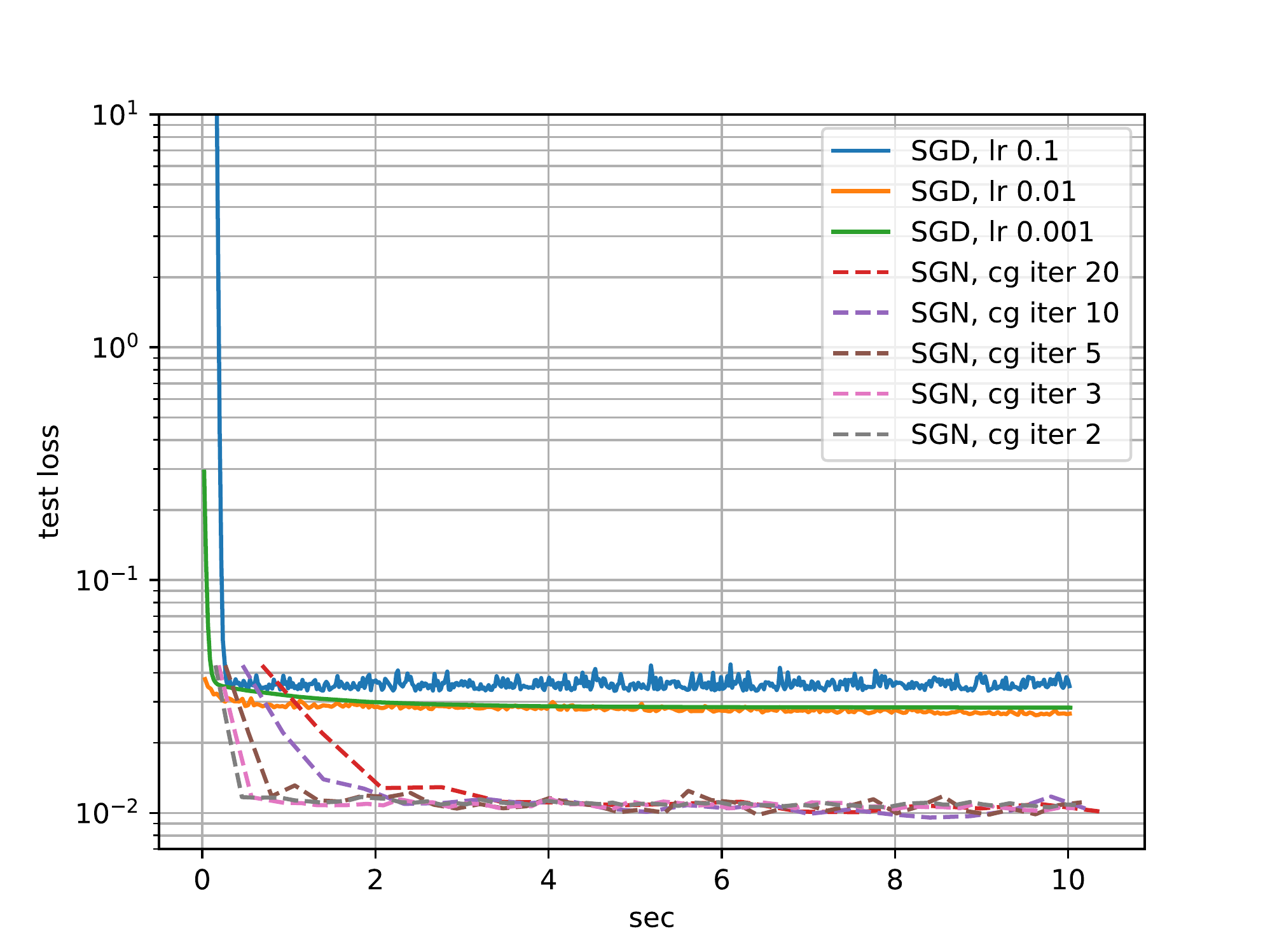}}
	
    \end{minipage}%
    \begin{minipage}{0.5\textwidth}
        {\includegraphics[width=1\textwidth]{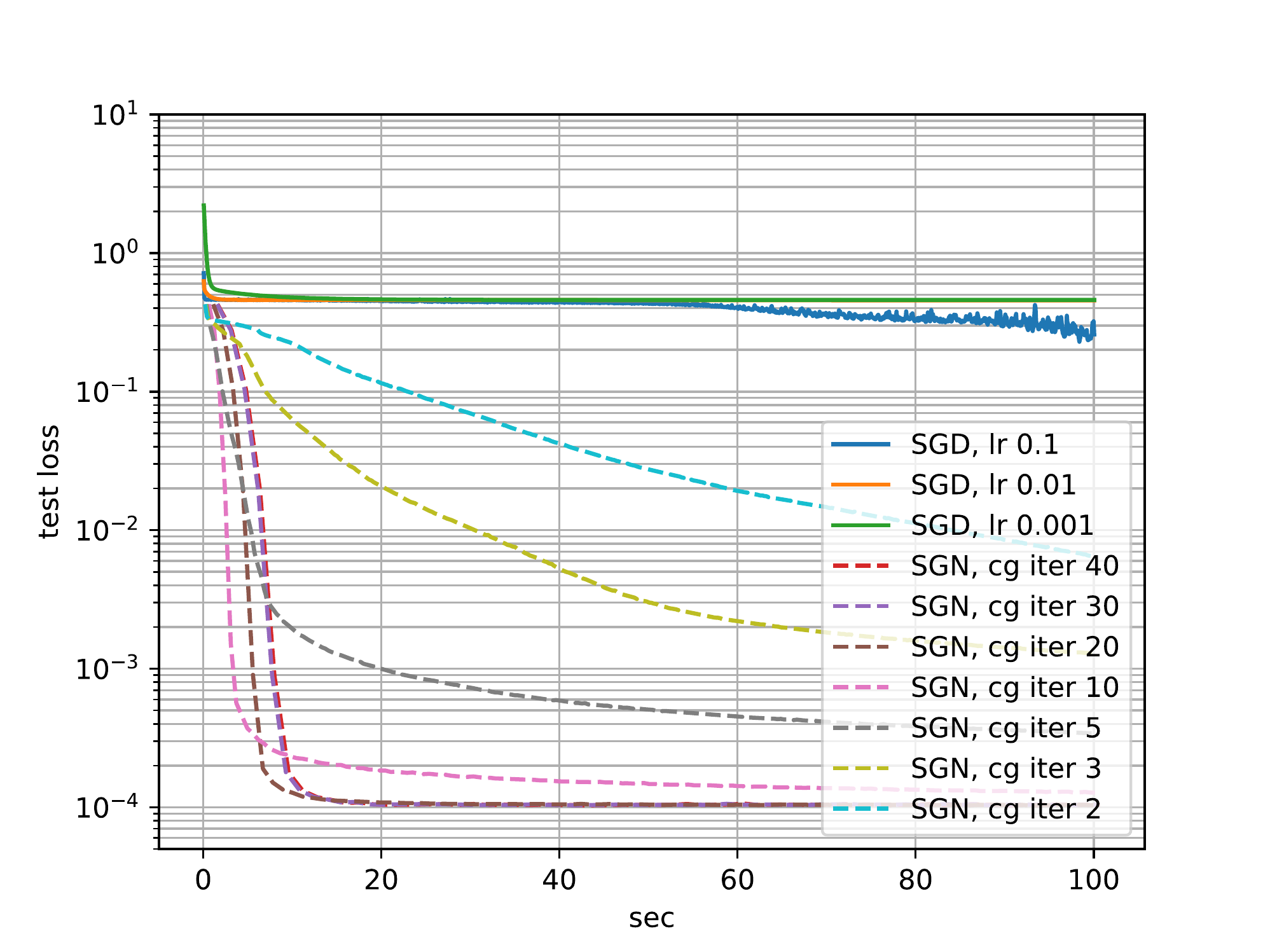}}

    \end{minipage}
    \caption{Train loss vs. seconds averaged over $5$ runs for SGN (dashed lines) and SGD (continuous lines) on boston housing (left) and the sine wave regression task (right). Different colors are used for different values of CG iterations and learning rate for SGN and SGD, respectively.}\label{fig:boston_sine_time}
\end{figure}

\subsection{Sine Wave Function}\label{subsec:sine}
We generate an artificial dataset of $10000$ samples with
$x \sim \mathcal{U}\left(-1, 1\right),\, y=\sin(10 \cdot x) + \mathcal{N}\left( 0,\, \sigma^2 \right)$ and $\sigma=0.01$.

As regression model we use a simple feedforward neural network with 3 hidden layers of 20 units each and sigmoid as activation function. We use the mean squared error as objective function and set the mini-batch size to $1000$ for both SGD and SGN. Similar conclusions as for the boston housing benchmarks can be drawn from the right plot in Figure~\ref{fig:boston_sine_time} and Figures~\ref{fig:sine_1}-\ref{fig:sine_4} in the Appendix. Regarding SGD, the results with learning rate value of 1 are not shown in the figures for clarity, as with this learning rate value SGD's iterates quickly diverge leading to high values of the objective function. The benchmarks are run on a Intel(R) Core(TM) i7-7560U CPU @ 2.40GHz.  

\subsection{MNIST}\label{subsec:mnist}
For this classification task we use the same architecture as in~\ref{subsec:sine}. Cross-entropy is used as objective function and the mini-batch size is set to $1000$ for both SGD and SGN. Table~\ref{tables:mnist}, together with the Figures~\ref{fig:mnist_1}-\ref{fig:mnist_4} in Section~\ref{sec:additional_figures} of the Appendix, summarizes the results of these benchmarks. SGN not only performs better than SGD, but also appears to be more robust to the number of CG iterations than SGD to the learning rate value. The benchmarks are run on a GeForce GTX TITAN X GPU. 
  \begin{table}[H]
    \begin{minipage}{.5\textwidth}
      \centering
      \begin{tabular}{lcccr}
\toprule
Algorithm &Test Accuracy $(\%)$& epochs \\
\midrule
SGD (0.001) & 17.2& 36\\
SGD (0.01) & 35.3& 36\\
SGD (0.1) & 71.9& 36\\
SGD (1) & 89.8& 36\\
SGD (10) & 11.6& 36\\
SGN (2)  & 91.6 & 10\\
\textbf{SGN (3)}  & \textbf{92.9} & \textbf{8}\\
SGN (5)  & 92.8 & 5\\
SGN (10)  & 91.6 & 3\\
\bottomrule
\end{tabular}

    \end{minipage}
    \begin{minipage}{.5\textwidth}
      \centering
      \begin{tabular}{lcccr}
\toprule
Algorithm &Test Accuracy $(\%)$& epochs \\
\midrule
SGD (0.001) & 25.4& 145\\
SGD (0.01) & 58.2& 145\\
SGD (0.1) & 85.5& 144\\
SGD (1) & 93.0& 144\\
SGD (10) & 93.0& 144\\
SGN (2)  & 93.8 & 43\\
\textbf{SGN (3)}  & \textbf{94.0} & \textbf{33}\\
SGN (5)  & 93.7 & 22\\
SGN (10)  & 91.5 & 12\\
\bottomrule
\end{tabular}
    \end{minipage}
\caption{Test accuracies after $25$ (left) and $100$ (right) seconds of training averaged over $5$ runs for SGN and SGD on MNIST. In parenthesis the value of learning rate and CG iterations used for SGD and SGN, respectively.}\label{tables:mnist}
  \end{table}

\subsection{FashionMNIST}
For this task, we use a convolutional neural network with a first layer of 6 $5\times 5$ filters with stride 1 followed by relu activation function and $2\times 2$ max-pooling with stride 2, and a second layer with $16$ $5\times 5$ filters with stride 1 followed by relu activation function and $2\times 2$ max-pooling with stride 2. The final classification part of the network is constituted by 2 feedforward layers of 120 and 84 units each with relu activation function. The mini-batch size is set to $1000$ for both SGD and SGN and cross-entropy is used as objective function. Table~\ref{tables:fashion} and the Figures~\ref{fig:fashion_1}-\ref{fig:fashion_4} in Section~\ref{sec:additional_figures} of the Appendix summarize the results of these benchmarks. SGN with different values for the number of CG iterations outperforms all run configurations of SGD, except SGD with learning rate of 0.1. We also run SGD with a learning rate of value 1 but with this configuration the method fails to achieve convergence. The benchmarks are run on a GeForce GTX TITAN X GPU.

\begin{table}[H]
    \begin{minipage}{.5\textwidth}
      \centering
\begin{tabular}{lcccr}
\toprule
Algorithm &Test Accuracy $(\%)$& epochs \\
\midrule
SGD (0.001) & 63.1& 15\\
SGD (0.01) & 82.3& 15\\
\textbf{SGD (0.1)} & \textbf{86.7}& \textbf{15}\\
SGN (2) & 84.5& 3\\
SGN (3) & 85.2 & 2\\
SGN (5)  & 85.6 & 1\\
SGN (10)  & 85.4 & 1\\
\bottomrule
\end{tabular}

    \end{minipage}
    \begin{minipage}{.5\textwidth}
      \centering
\begin{tabular}{lcccr}
\toprule
Algorithm &Test Accuracy $(\%)$& epochs \\
\midrule
SGD (0.001) & 76.5& 46\\
SGD (0.01) & 85.5& 46\\
\textbf{SGD (0.1)} & \textbf{88.5}& \textbf{46}\\
SGN (2) & 86.2& 10\\
SGN (3) & 87.2 & 8\\
SGN (5)  & 87.8 & 5\\
SGN (10)  & 86.0 & 3\\
\bottomrule
\end{tabular}
    \end{minipage}
\caption{Test accuracies after $50$ (left) and $150$ (right) seconds of training averaged over $5$ runs for SGN and SGD on FashionMNIST. In parenthesis the value of learning rate and CG iterations used for SGD and SGN, respectively.}\label{tables:fashion}
  \end{table}

\subsection{CIFAR10}
For the CIFAR10-classification task we use a convolutional neural network with a first layer of 32 $3\times 3$ filters with stride 1 followed by relu activation function and $2\times 2$ max-pooling with stride 2, a second layer with $64$ $3\times 3$ filters with stride 1 followed by relu activation function and $2\times 2$ max-pooling with stride 2, and a final third layer with $64$ $3\times 3$ filters with stride 1 followed by relu activation function. The final classification part of the network is constituted by 1 feedforward layer of 64 units with relu activation function. The mini-batch size is set to $1000$ for both SGD and SGN and cross-entropy is used as objective function. Table~\ref{tables:cifar10}, together with Figures~\ref{fig:cifar10_1}-\ref{fig:cifar10_8} in Section~\ref{sec:additional_figures} of the Appendix, summarizes the performance of SGD and SGN on this task for different values of learning rate and number of CG iterations, respectively.

\begin{table}[H]
    \begin{minipage}{.5\textwidth}
      \centering
\begin{tabular}{lcccr}
\toprule
Algorithm &Test Accuracy $(\%)$& epochs \\
\midrule
SGD (0.001) & 10.0& 46\\
SGD (0.01) & 19.0& 45\\
SGD (0.1) & 63.1 & 45\\
SGN (2) & 44.9 & 10\\
SGN (3) & 49.3 & 8\\
SGN (5)  & 56.9 & 5\\
\textbf{SGN (10)} & \textbf{64.7} & \textbf{3}\\
\bottomrule
\end{tabular}

    \end{minipage}
    \begin{minipage}{.5\textwidth}
      \centering
\begin{tabular}{lcccr}
\toprule
Algorithm &Test Accuracy $(\%)$& epochs \\
\midrule
SGD (0.1) & 60.9 & 135\\
\textbf{SGN (10)} & \textbf{71.0} & \textbf{9}\\
\bottomrule
\end{tabular}
    \end{minipage}
\caption{Test accuracies after $650$ (left) and $2000$ (right) seconds of training averaged over $5$ runs for SGN and SGD on CIFAR10. In parenthesis the value of learning rate and CG iterations used for SGD and SGN, respectively.}\label{tables:cifar10}
  \end{table}
\subsection{General and Qualitative Considerations}
In order for the stochastic Gauss-Newton Hessian approximation to be significant for the target problem, a bigger size of mini-batch is generally needed. The number of CG iterations is highly related to the size of the mini-batch as well as the size of the residuals: if the stochastic Gauss-Newton Hessian approximation is not ``reliable'' because the mini-batch is not (enough) representative of the dataset and/or the residuals are ``big'', then it is generally better to perform a smaller number of CG iterations since the direction provided by the exact solution of the linear system~\eqref{eq: lin_system} is probably not a good direction for the target problem~\eqref{eq: gen_problem}. On the other hand, if the mini-batch is (enough) representative of the dataset, the stochastic Gauss-Newton Hessian approximation, given that the residuals are small, is a good and ``reliable'' approximation of the Hessian matrix, and therefore we can and should perform more CG iterations to fully take advantage of the curvature information. In general, the less CG iterations we perform, the less expensive are the SGN's iterations and the more SGN is similar to SGD (in the limit, i.e. only one CG iteration, the two methods are equivalent). 
In our benchmarks we can observe that, in the big mini-batch scenario, the performance of SGN is quite robust with respect to the number of CG iterations. Assuming that the residuals are ``small enough'', as an heuristic for the big mini-batch scenario we suggest to set the CG iterations to the highest value for which SGN's performance in terms of time is still competitive with that of SGD. 
Another interesting direction could also be to study adaptive heuristics to adjust the number of CG iterations during the curse of the optimization. For instance, a trust-region scheme could be adopted for adaptively adjusting the number of CG iterations as our Theano implementation currently offers as feature for the Levenberg-Marquardt parameter.

Despite the higher cost of its iterations, SGN shows to be very competitive against SGD across all the benchmarks. The better convergence rate of SGN with enough CG iterations clearly appears when we compare the performance of SGN and SGD in terms of number of epochs (see for instance Figure~\ref{fig:cifar10_2} in Section~\ref{sec:additional_figures} of the Appendix as well as the test accuracy results with the corresponding number of epochs for both SGN and SGD in the Tables~\ref{tables:mnist}-\ref{tables:cifar10}). Being the iterations significantly more expensive than that of SGD for reasonable values of CG iterations, it is hence important to consider the timings, where the better convergence rate does not always appear clearly. Nevertheless, SGN remains competitive also when we look at the performance in terms of time, and it generally leads to solutions with better generalization properties. 

Another really desirable aspect of SGN that emerges from our benchmarks is its robustness against its hyperparameters, while the performance of SGD appears to be very sensitive to the value of the learning rate, e.g. leading to convergence to minimizers with dramatically different generalization properties, slowing down the convergence rate or even preventing the iterates from convergence.  

\noindent{\textbf{Conclusions}}
With this work we mathematically described, implemented and benchmarked SGN: a stochastic second-order and Hessian-free method based on the Gauss-Newton Hessian approximation for training deep neural networks. 
Although SGN as optimization algorithm for training deep neural networks was already introduced back in 2010 by Martens~\cite{icml2010_094}, to the best of our knowledge, ours is the first empirical study of its performance and available efficient implementation in a popular deep learning framework.  
As already discussed, there are a lot of interesting and promising future directions to extend this work such as the development of an adaptive scheme to adjust the number of CG iterations and a more extensive benchmarking on networks with state-of-the-art architectures. 
\section*{Broader Impact}
If on one hand, in order for second-order methods to work efficiently on large-scale scenarios ad-hoc implementations are required, on the other hand, they might lead to significant energy savings since, as shown in our experiments, they are generally more robust against their hyperparameters than first-order methods. The possibility of shortening or even avoiding the expensive process of hyperparameter tuning comes together with a competitive runtime, which also goes in the direction of a more sustainable AI.

We hope that our work will contribute to increase the curiosity of the deep learning community into the exploration of more sophisticated optimization methods, which is to some extent in contrast with the major tendency of engineering the network architectures for SGD, e.g. by using over-parametrization or adding skip connections. To ease such exploration, we hope more sophisticated optimization features, such as forward automatic differentiation, will soon be officially integrated in all deep learning frameworks, especially in Pytorch and Tensorflow. 

Finally, the authors believe that this work does not present any foreseeable negative societal consequence.




\begin{ack}
This work has partly been supported by the European Research Council (ERC) under the European Union’s Horizon 2020 research and innovation programme under grant no.\ 716721 as well as by the German Federal Ministry for Economic Affairs and Energy (BMWi) via DyConPV (0324166B), and by DFG via Research Unit FOR 2401. The authors thank James Martens for the insightful discussions about Hessian-free methods.
\end{ack}
\bibliographystyle{plain}
\bibliography{neurips_2020}
\appendix
\onecolumn
\section{Algorithms}\label{sec:algorithms}
 
\begin{algorithm}[h]
   \caption{SGN for~\eqref{eq:prob2}}
   \label{alg:example}
\begin{algorithmic}
   \STATE {\bfseries Input:} $\boldsymbol{\theta}_0\,,\,k_\mathrm{max}>0\,,\,\rho >0$
   \STATE {\bfseries Set:} $t\leftarrow 0$
	\REPEAT   
   \STATE  $J_{L}(\boldsymbol{\theta}_t)$ given by eq.~\eqref{eq:jacobian} (compute with backpropagation)
   \vspace{0.4mm}
   \STATE  $H_{L}^{GN}(\boldsymbol{\theta}_t)$ given by eq.~\eqref{eq:LMGN_hessian} (compute with forward and reverse AD)
   \STATE {\bfseries Call to algorithm~\ref{alg:CG_alg}:} $\Delta{\boldsymbol{\theta}}_{t+1}$ 
   \STATE $\boldsymbol{\theta}_{t+1}\leftarrow \boldsymbol{\theta}_{t} +  \Delta\boldsymbol{\theta}_{t+1}$
   \STATE $t \leftarrow t+1$
   \UNTIL convergence conditions hold
\end{algorithmic}
\end{algorithm}
\vspace{2cm}
\begin{algorithm}[h]
   \caption{CG for~\eqref{eq:linear_system}}
   \label{alg:CG_alg}
\begin{algorithmic}
   \STATE {\bfseries Input:} $\Delta\boldsymbol{\theta}_0$, $k_\mathrm{max}>0$
   \STATE {\bfseries Set: } $\mathbf{r}_0 \leftarrow \left( H_L^{GN}+\rho I \right)\Delta\boldsymbol{\theta}_0 + J_L^\top$
   \STATE {\bfseries Set: } $\mathbf{p}_0 \leftarrow -\mathbf{r}_0$
   \STATE {\bfseries Set: } $k \leftarrow 0$
   \REPEAT
   \STATE $\alpha_k\leftarrow \frac{\mathbf{r}_k^\top\mathbf{r}_k}{\mathbf{p}_k^\top \left( H_L^{GN}+\rho I \right)\mathbf{p}_k}$
   \STATE $\Delta\boldsymbol{\theta}_{k+1}\leftarrow \Delta\boldsymbol{\theta}_k + \alpha_k \mathbf{p}_k$
   \STATE $\mathbf{r}_{k+1}\leftarrow \mathbf{r}_k + \alpha_k \left( H_L^{GN}+\rho I \right)\mathbf{p}_k$
   \STATE $\beta_k\leftarrow \frac{\mathbf{r}_{k+1}^\top\mathbf{r}_{k+1}}{\mathbf{r}_{k}^\top\mathbf{r}_{k}}$
   \STATE $\mathbf{p}_{k+1}\leftarrow -\mathbf{r}_{k+1} +\beta_k\mathbf{p}_k$
   \STATE $k\leftarrow k+1$   
   \UNTIL{$k<k_\mathrm{max}$} or convergence conditions hold
   \STATE {\bfseries Return:} $\Delta\boldsymbol{\theta}_k$
\end{algorithmic}
\end{algorithm}

\newpage
\section{Additional Figures}\label{sec:additional_figures}
\begin{figure}[!htb]
    \centering
    \begin{minipage}{.8\textwidth}
        \centering
        {\includegraphics[width=\textwidth]{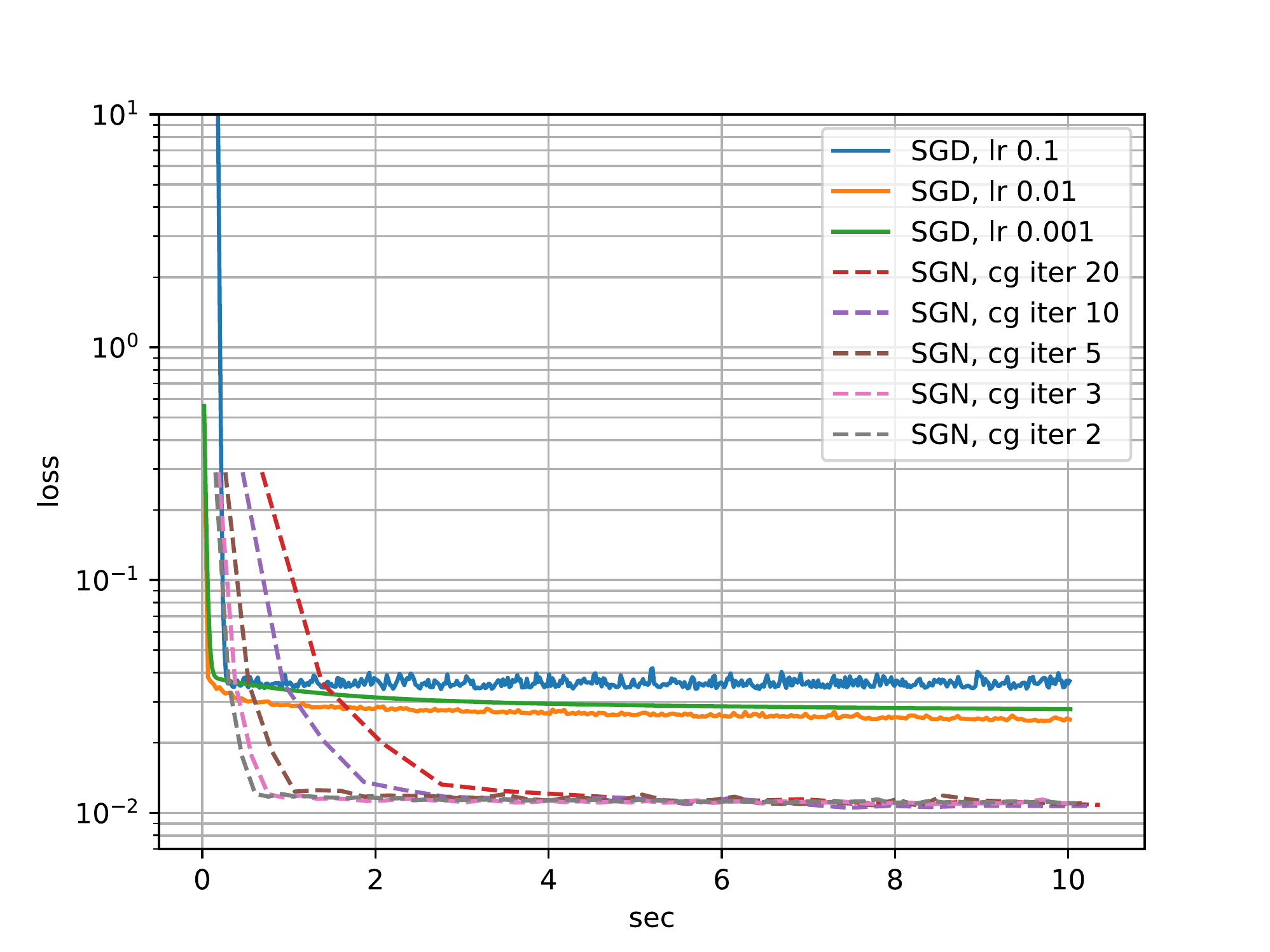}}
\caption{Average train loss vs. seconds over $5$ runs for SGN (dashed lines) and SGD (continuous lines) on boston housing. Different colors are used for different values of CG iterations and learning rate for SGN and SGD respectively. A batch size of $101$ is used for both SGD and SGN.}\label{fig:boston_1}
    \end{minipage}\qquad
    \begin{minipage}{0.8\textwidth}
         {\includegraphics[width=\textwidth]{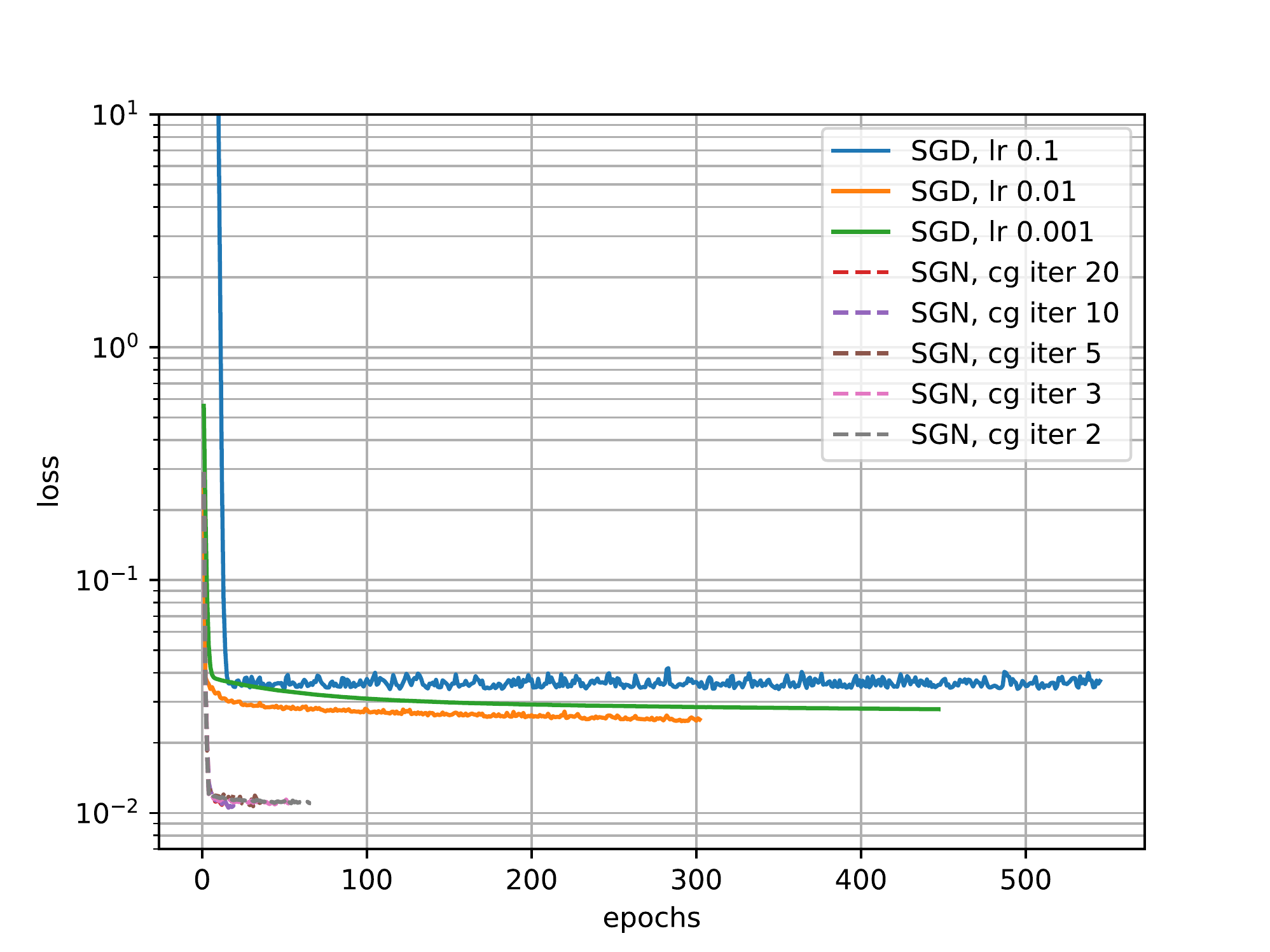}}
\caption{Average train loss vs. epochs over $5$ runs for SGN (dashed lines) and SGD (continuous lines) on boston housing. Different colors are used for different values of CG iterations and learning rate for SGN and SGD respectively. A batch size of $101$ is used for both SGD and SGN.}\label{fig:boston_2}
    \end{minipage}
\end{figure}

\begin{figure}[!htb]
    \centering
    \begin{minipage}{.8\textwidth}
        \centering
        {\includegraphics[width=\textwidth]{./figures/boston_testloss_time.pdf}}
\caption{Average test loss vs. seconds over $5$ runs for SGN (dashed lines) and SGD (continuous lines) on boston housing. Different colors are used for different values of CG iterations and learning rate for SGN and SGD respectively. A batch size of $101$ is used for both SGD and SGN.}\label{fig:boston_3}
    \end{minipage}\qquad
        \begin{minipage}{0.8\textwidth}
         {\includegraphics[width=\textwidth]{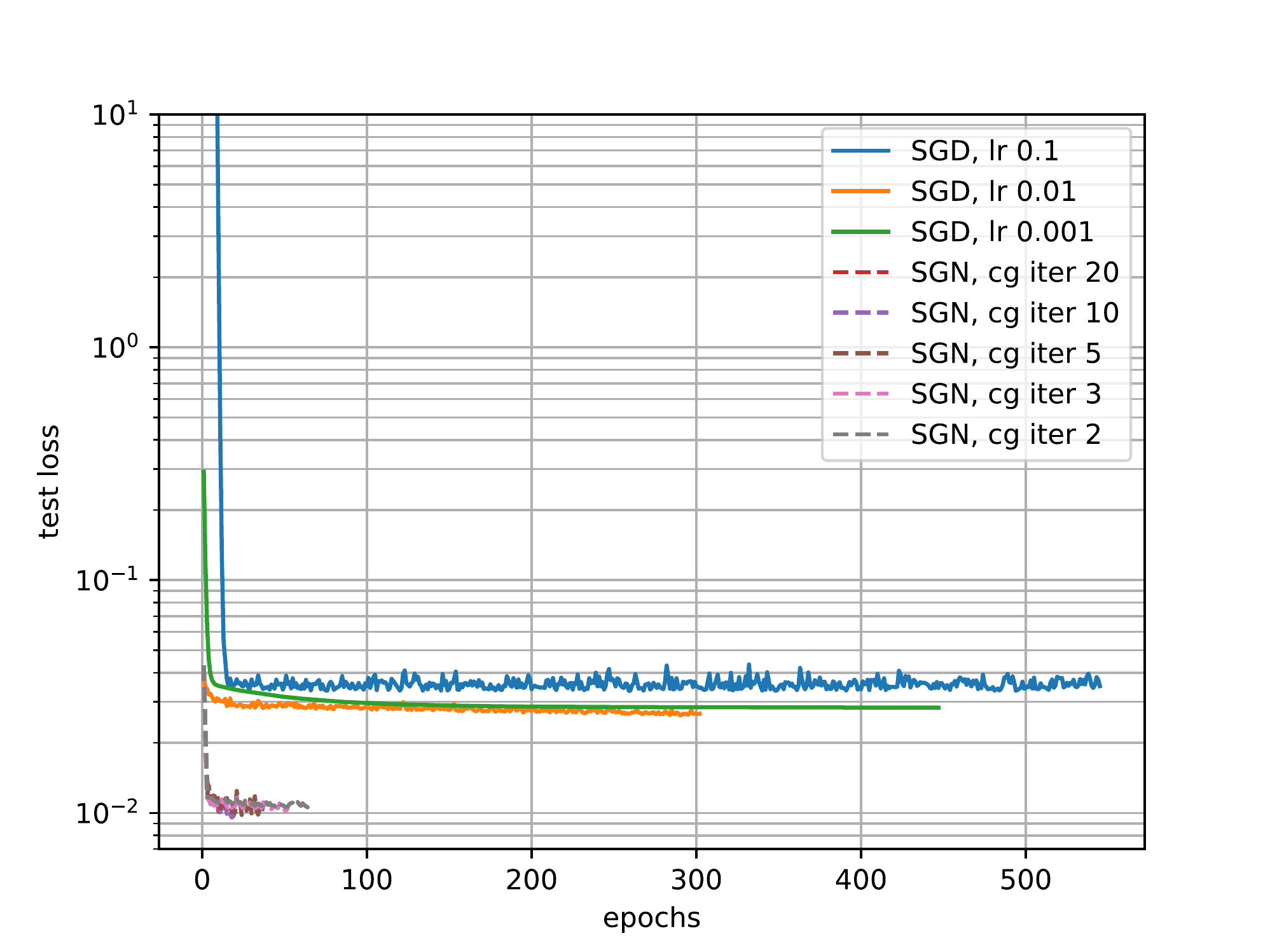}}
\caption{Average test loss vs. epochs over $5$ runs for SGN (dashed lines) and SGD (continuous lines) on boston housing. Different colors are used for different values of CG iterations and learning rate for SGN and SGD respectively. A batch size of $101$ is used for both SGD and SGN.}\label{fig:boston_4}
    \end{minipage}
\end{figure}

\newpage

\begin{figure}[!htb]
    \centering
    \begin{minipage}{.8\textwidth}
        \centering
        {\includegraphics[width=\textwidth]{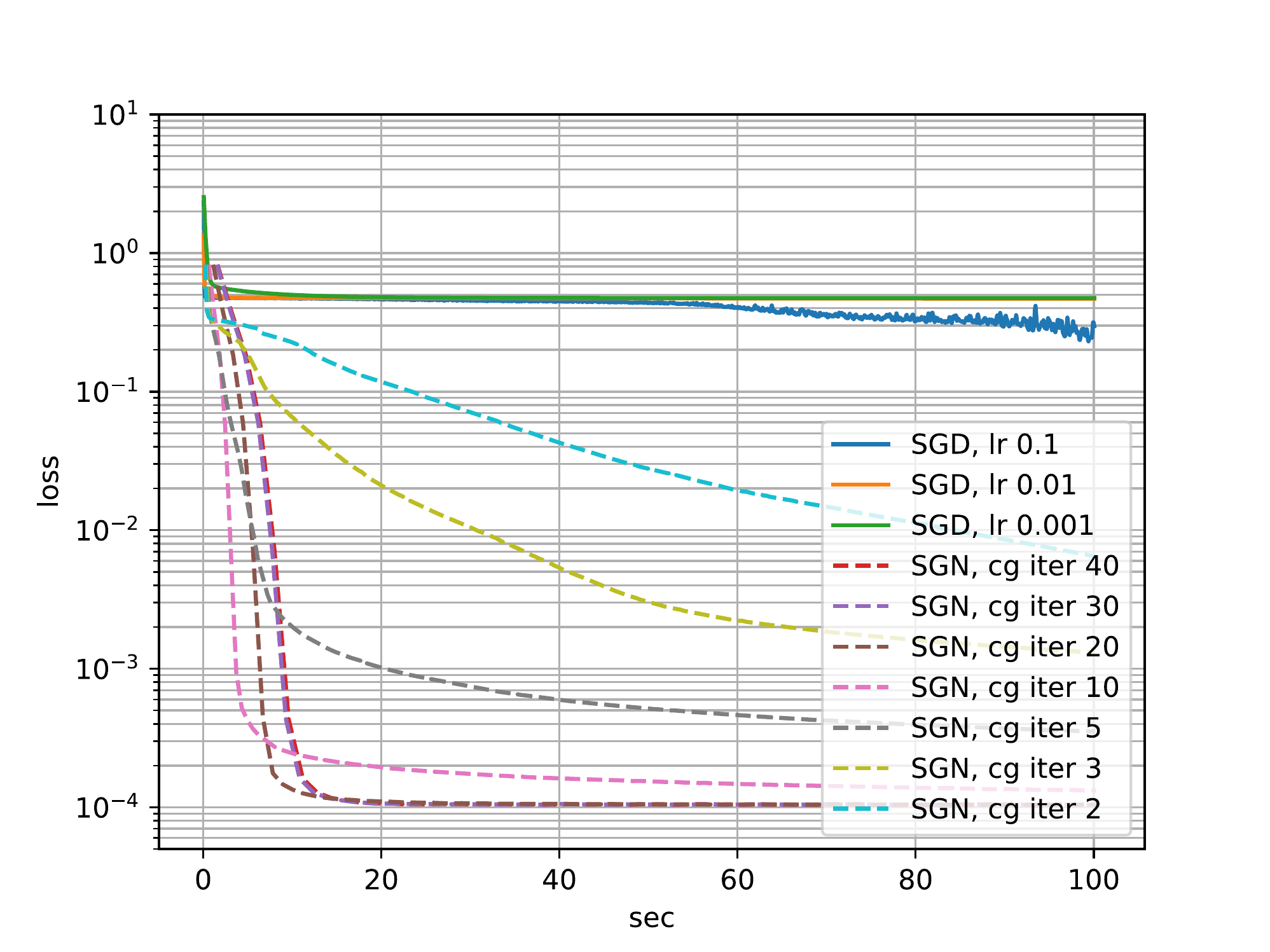}}
\caption{Average train loss vs. seconds over $5$ runs for SGN (dashed lines) and SGD (continuous lines) for the noisy sine wave function regression task. Different colors are used for different values of CG iterations and learning rate for SGN and SGD respectively. A batch size of $5$ is used for both SGD and SGN.}\label{fig:sine_1}
    \end{minipage}\qquad
    \begin{minipage}{0.8\textwidth}
         {\includegraphics[width=\textwidth]{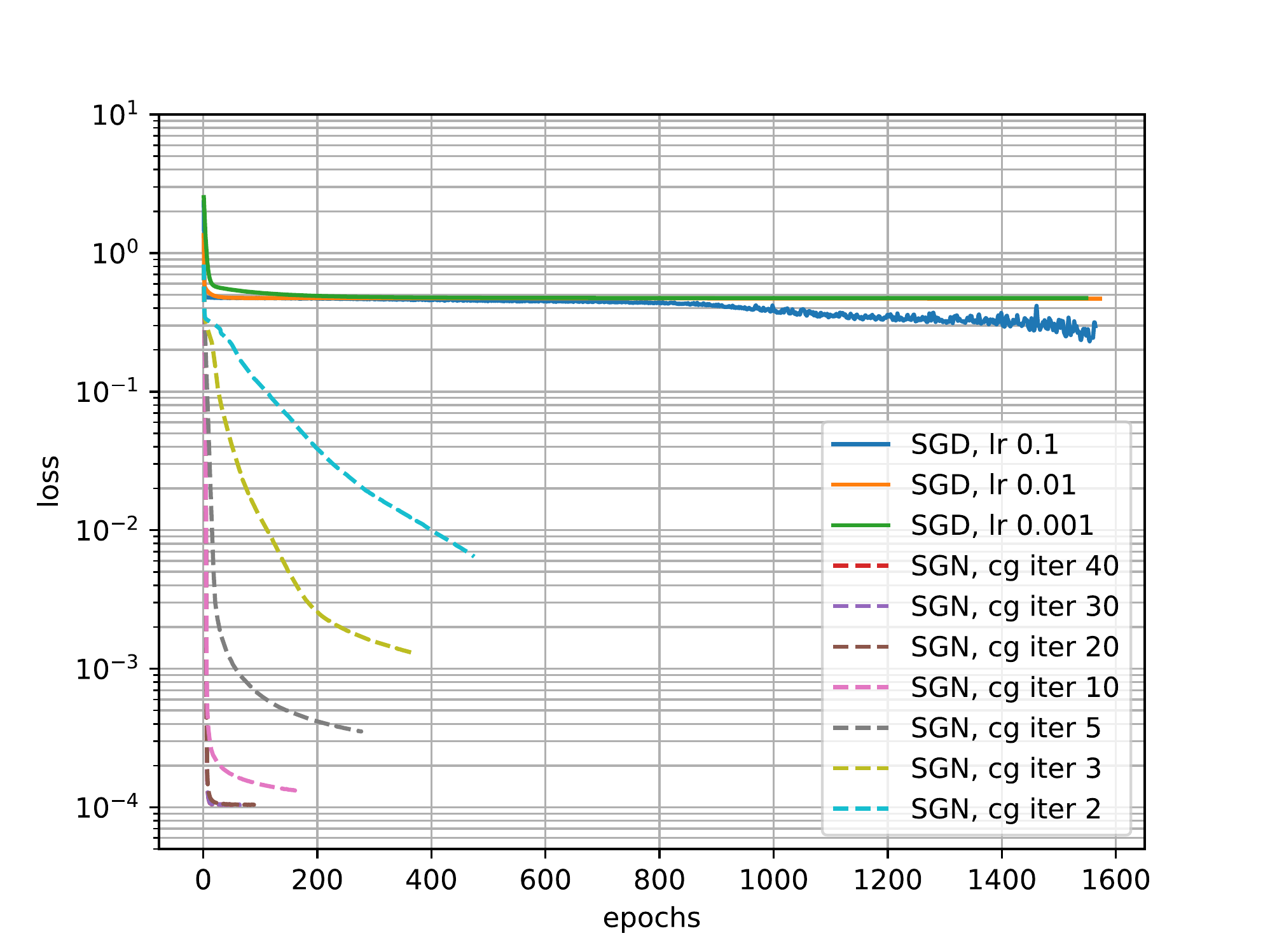}}
\caption{Average train loss vs. epochs over $5$ runs for SGN (dashed lines) and SGD (continuous lines) for the noisy sine wave function regression task. Different colors are used for different values of CG iterations and learning rate for SGN and SGD respectively. A batch size of $5$ is used for both SGD and SGN.}\label{fig:sine_2}
    \end{minipage}
\end{figure}

\begin{figure}[!htb]
    \centering
    \begin{minipage}{.8\textwidth}
        \centering
        {\includegraphics[width=\textwidth]{./figures/sine10_testloss_time.pdf}}
\caption{Average test loss vs. seconds over $5$ runs for SGN (dashed lines) and SGD (continuous lines) for the noisy sine wave function regression task. Different colors are used for different values of CG iterations and learning rate for SGN and SGD respectively. A batch size of $5$ is used for both SGD and SGN.}\label{fig:sine_3}
    \end{minipage}\qquad
        \begin{minipage}{0.8\textwidth}
         {\includegraphics[width=\textwidth]{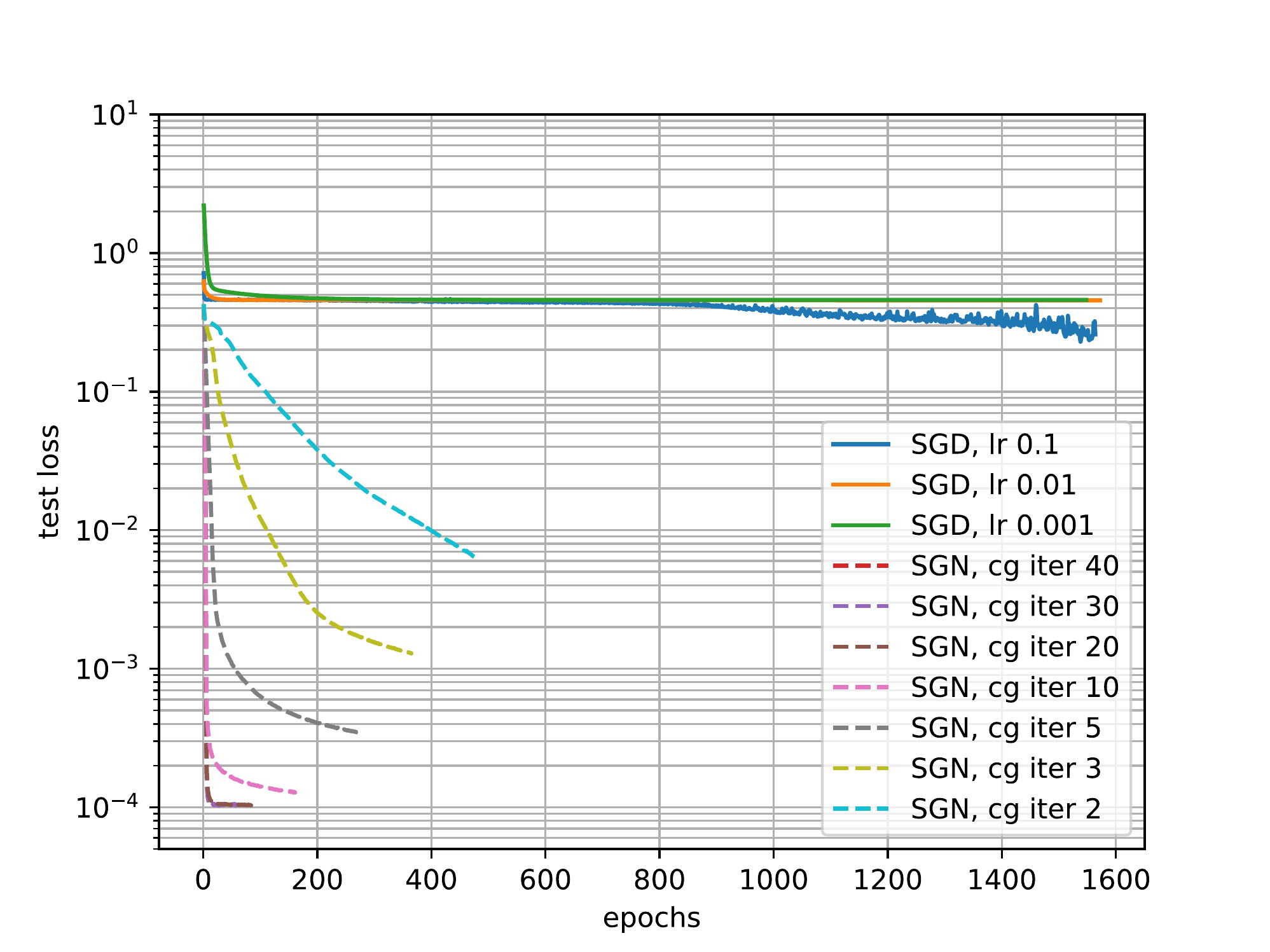}}
\caption{Average test loss vs. epochs over $5$ runs for SGN (dashed lines) and SGD (continuous lines) for the noisy sine wave function regression task. Different colors are used for different values of CG iterations and learning rate for SGN and SGD respectively. A batch size of $5$ is used for both SGD and SGN.}\label{fig:sine_4}
    \end{minipage}
\end{figure}

\newpage
\begin{figure}[!htb]
    \centering
    \begin{minipage}{.8\textwidth}
        \centering
       {\includegraphics[width=\textwidth]{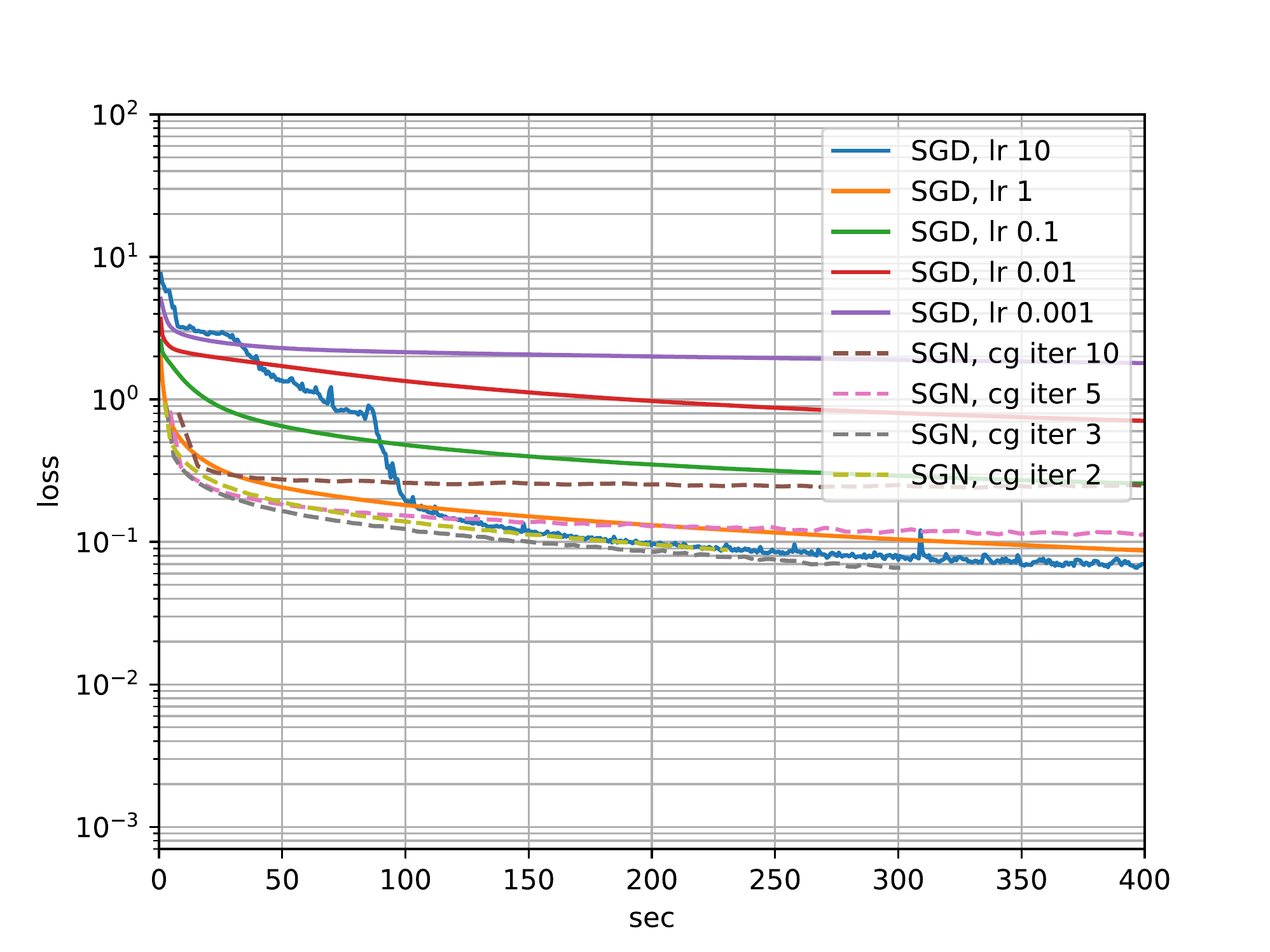}}
\caption{Average train loss vs. seconds over $5$ runs for SGN (dashed lines) and SGD (continuous lines) on MNIST. Different colors are used for different values of CG iterations and learning rate for SGN and SGD respectively. A batch size of $1000$ is used for both SGD and SGN.}\label{fig:mnist_1}
    \end{minipage}\qquad
    \begin{minipage}{.8\textwidth}
        \centering
       {\includegraphics[width=\textwidth]{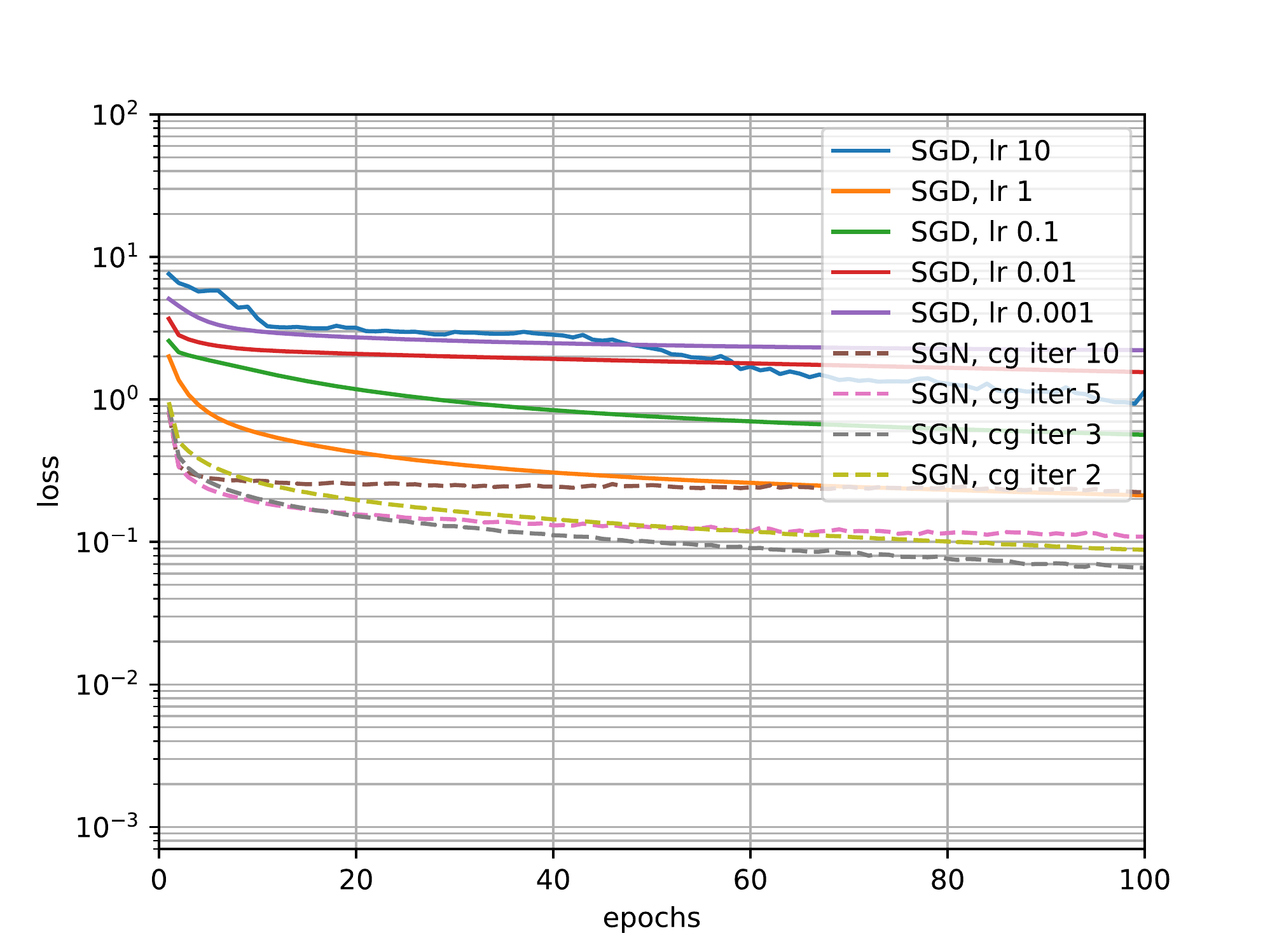}}
\caption{Average train loss vs. epochs over $5$ runs for SGN (dashed lines) and SGD (continuous lines) on MNIST. Different colors are used for different values of CG iterations and learning rate for SGN and SGD respectively. A batch size of $1000$ is used for both SGD and SGN.}\label{fig:mnist_2}
    \end{minipage}
\end{figure}
\begin{figure}[!htb]
    \centering
    \begin{minipage}{.8\textwidth}
        \centering
       {\includegraphics[width=\textwidth]{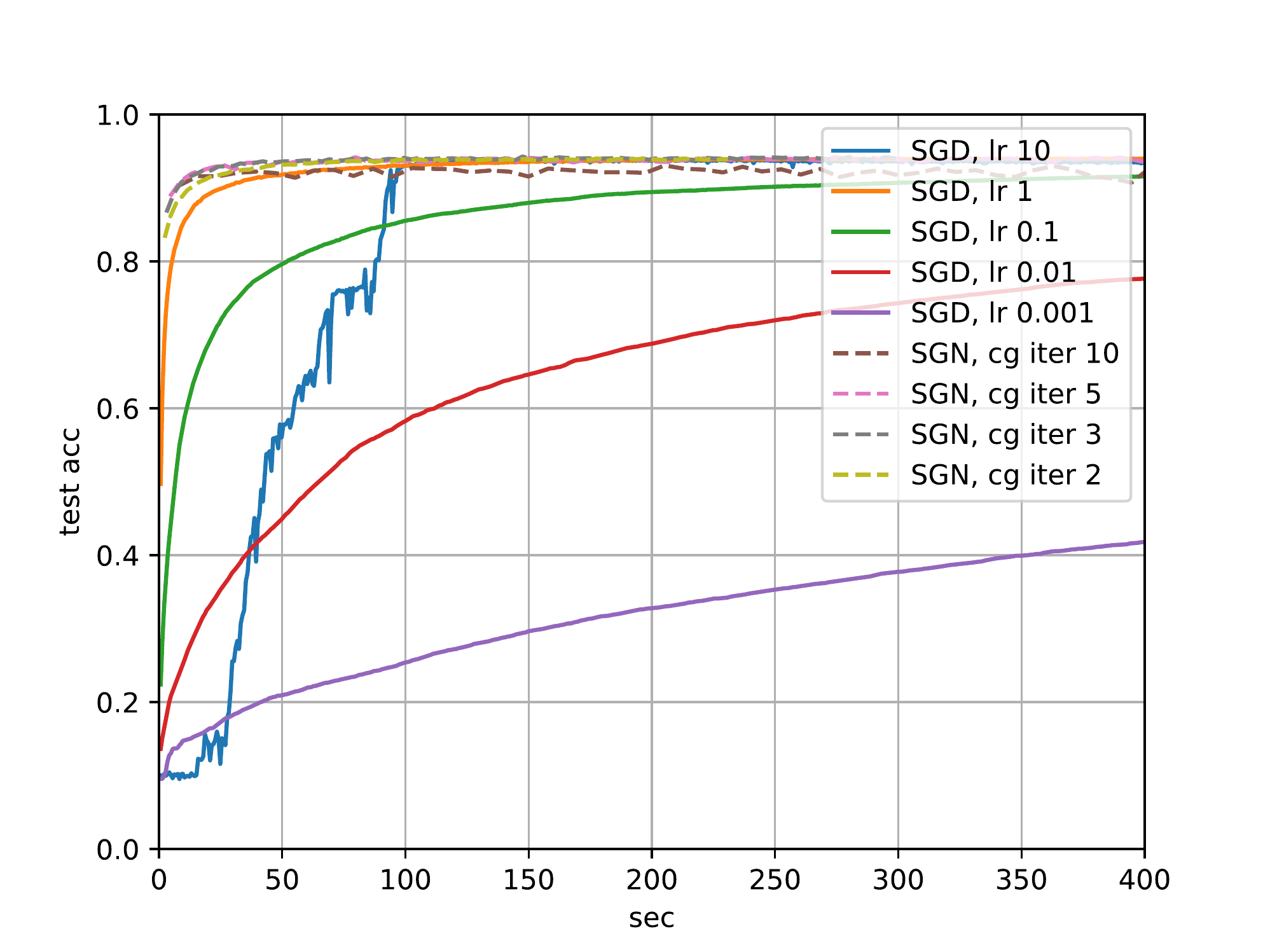}}
\caption{Average test accuracy vs. seconds over $5$ runs for SGN (dashed lines) and SGD (continuous lines) on MNIST. Different colors are used for different values of CG iterations and learning rate for SGN and SGD respectively. A batch size of $1000$ is used for both SGD and SGN.}\label{fig:mnist_3}
    \end{minipage}\qquad
    \begin{minipage}{0.8\textwidth}
        \centering
       {\includegraphics[width=\textwidth]{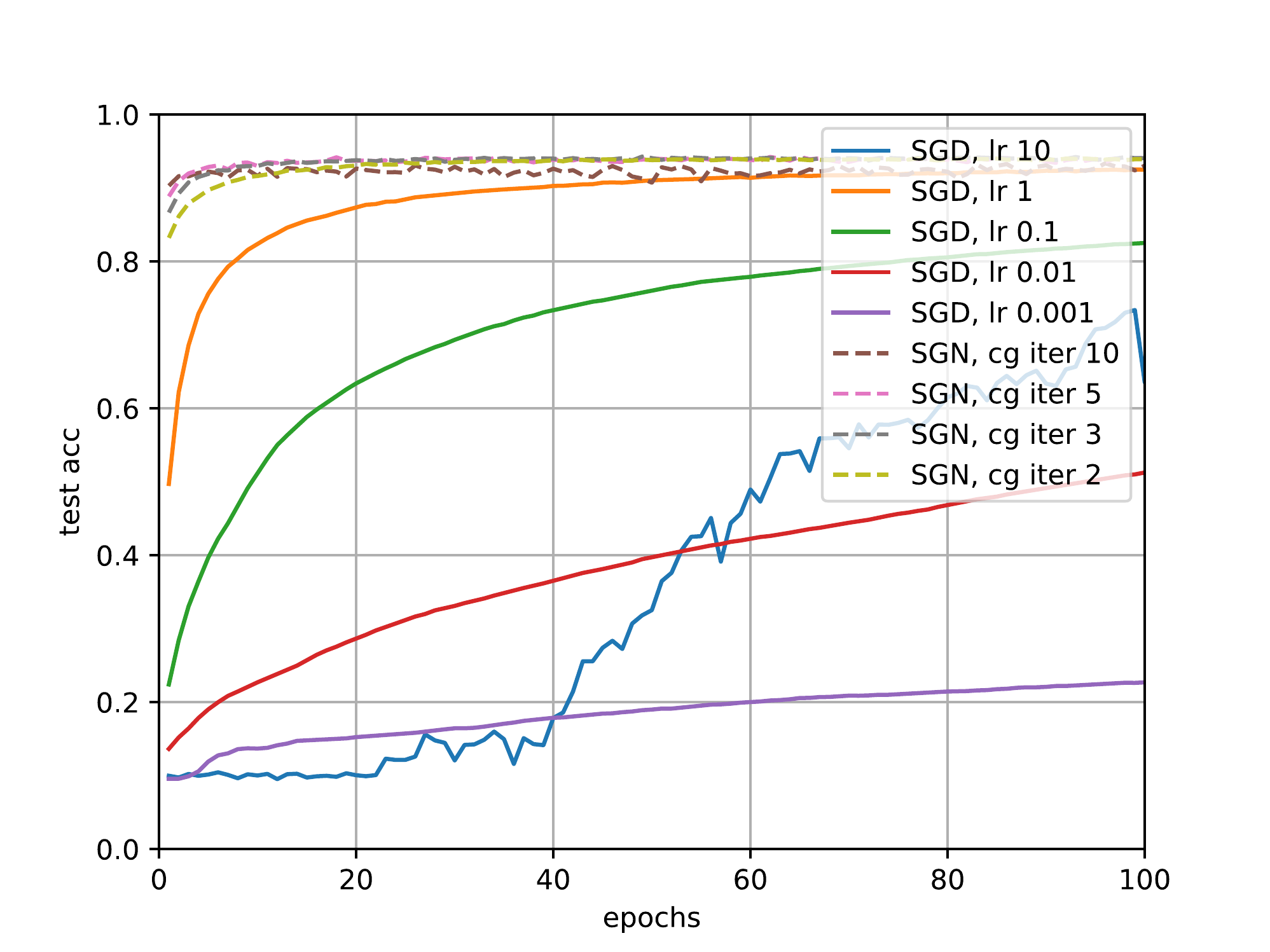}}
\caption{Average test accuracy vs. epochs over $5$ runs for SGN (dashed lines) and SGD (continuous lines) on MNIST. Different colors are used for different values of CG iterations and learning rate for SGN and SGD respectively. A batch size of $1000$ is used for both SGD and SGN.}\label{fig:mnist_4}
    \end{minipage}
\end{figure}
\newpage
\begin{figure}[!htb]
    \centering
    \begin{minipage}{.8\textwidth}
        \centering
       {\includegraphics[width=\textwidth]{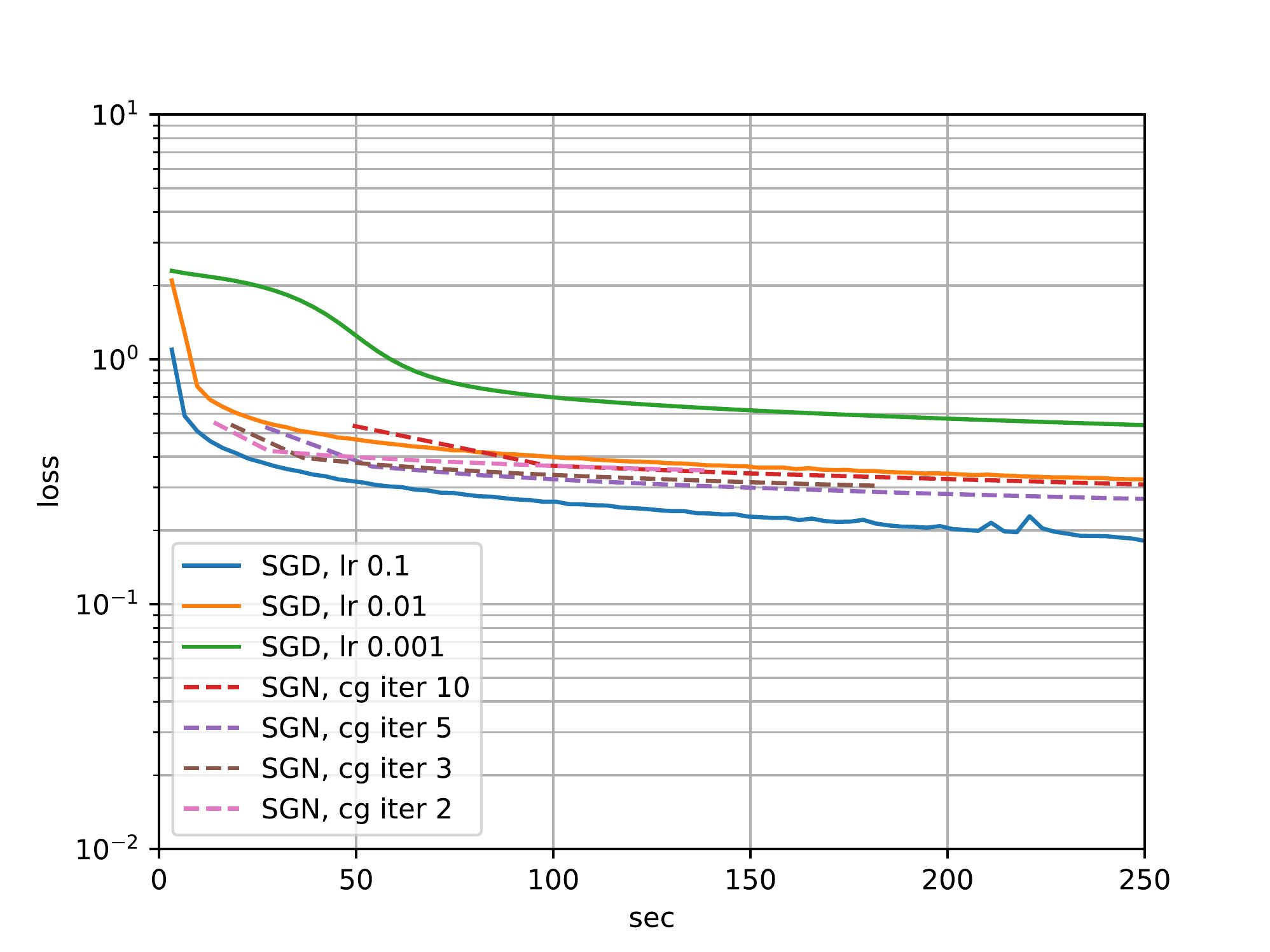}}
\caption{Average train loss vs. seconds over $5$ runs for SGN (dashed lines) and SGD (continuous lines) on FashionMNIST. Different colors are used for different values of CG iterations and learning rate for SGN and SGD respectively. A batch size of $1000$ is used for both SGD and SGN.}\label{fig:fashion_1}
    \end{minipage}\qquad
    \begin{minipage}{0.8\textwidth}
        \centering
        {\includegraphics[width=\textwidth]{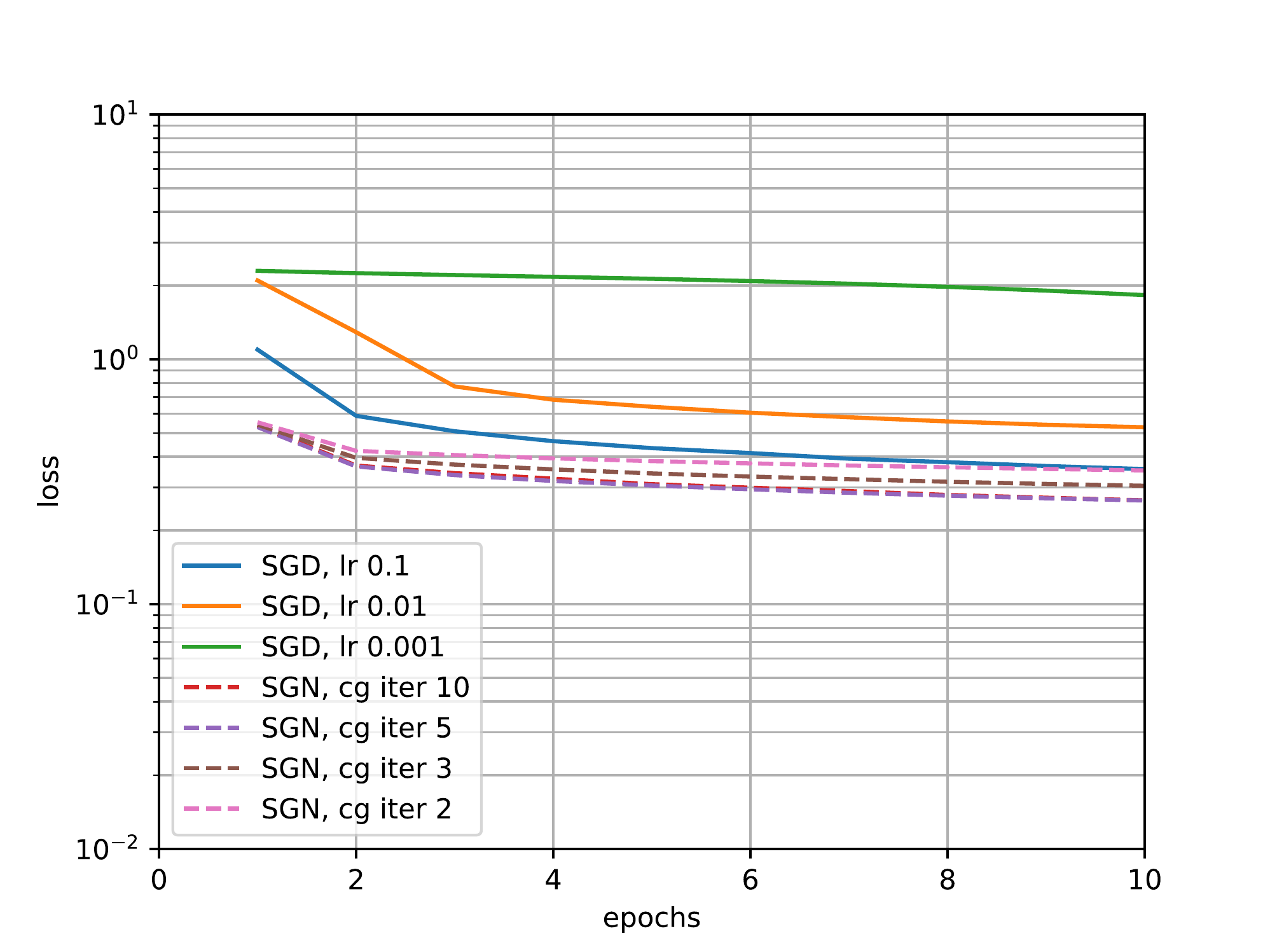}}
\caption{Average train loss vs. epochs over $5$ runs for SGN (dashed lines) and SGD (continuous lines) on FashionMNIST. Different colors are used for different values of CG iterations and learning rate for SGN and SGD respectively. A batch size of $1000$ is used for both SGD and SGN.}\label{fig:fashion_2}
    \end{minipage}
\end{figure}
\begin{figure}[!htb]
    \centering
    \begin{minipage}{.8\textwidth}
        \centering
       {\includegraphics[width=\textwidth]{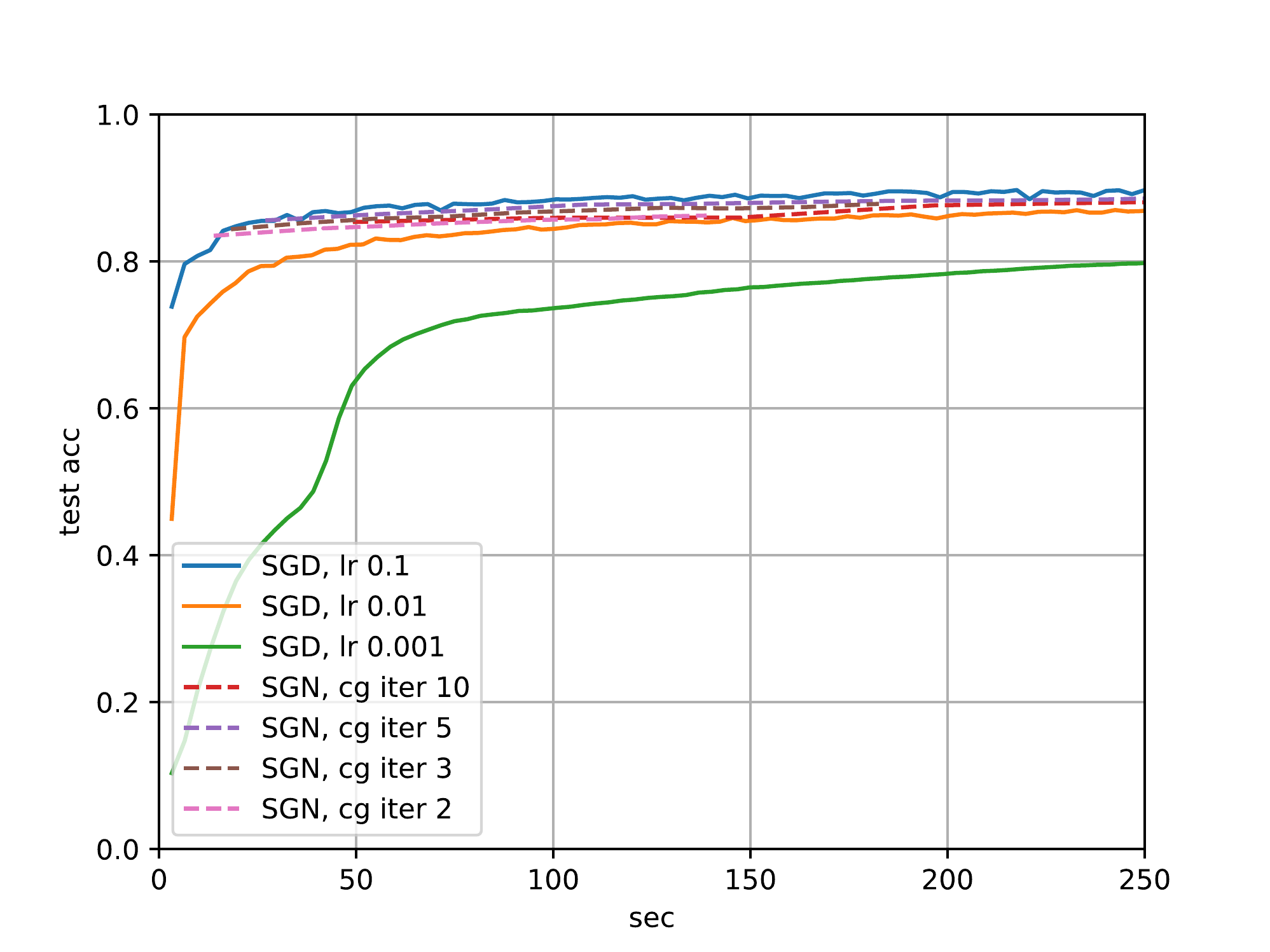}}
\caption{Average train loss vs. seconds over $5$ runs for SGN (dashed lines) and SGD (continuous lines) on FashionMNIST. Different colors are used for different values of CG iterations and learning rate for SGN and SGD respectively. A batch size of $1000$ is used for both SGD and SGN.}\label{fig:fashion_3}
    \end{minipage}\qquad
    \begin{minipage}{0.8\textwidth}
        \centering
        {\includegraphics[width=\textwidth]{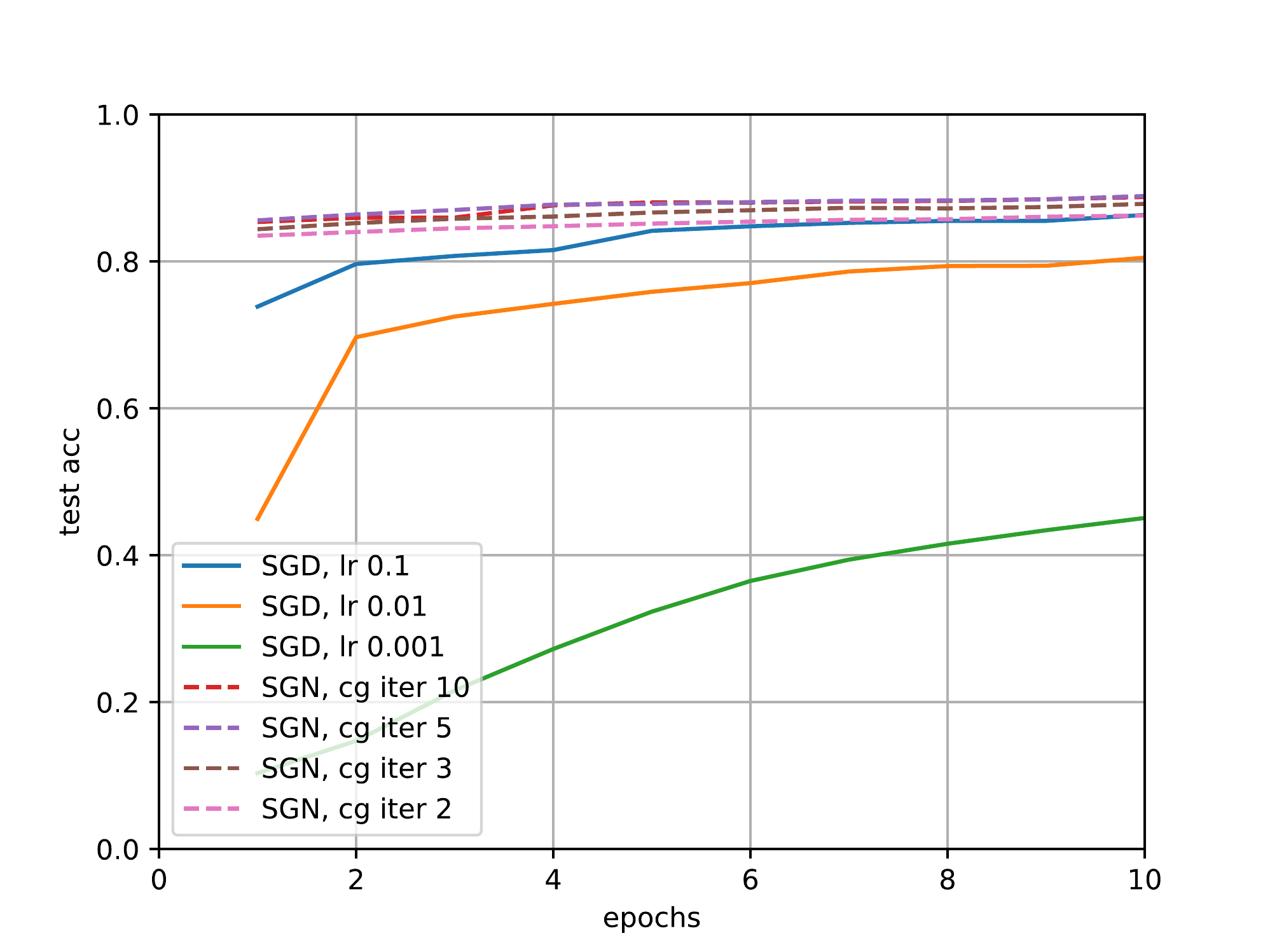}}
\caption{Average train loss vs. epochs over $5$ runs for SGN (dashed lines) and SGD (continuous lines) on FashionMNIST. Different colors are used for different values of CG iterations and learning rate for SGN and SGD respectively. A batch size of $1000$ is used for both SGD and SGN.}\label{fig:fashion_4}
    \end{minipage}
\end{figure}

\newpage
\begin{figure}[!htb]
    \centering
    \begin{minipage}{.8\textwidth}
        \centering
        {\includegraphics[width=\textwidth]{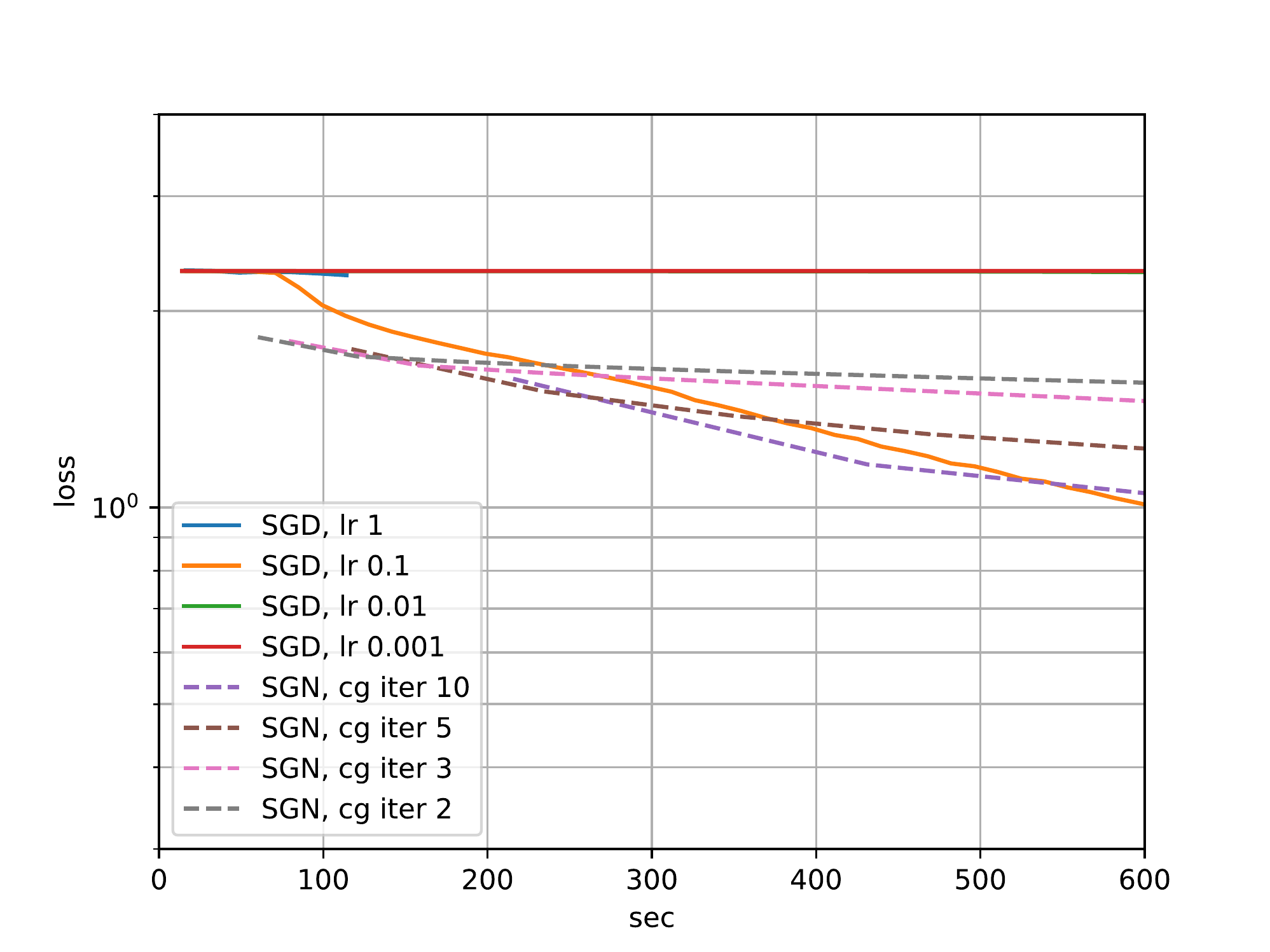}}
\caption{Average train loss vs. seconds over $5$ runs for SGN (dashed lines) and SGD (continuous lines) on CIFAR10. Different colors are used for different values of CG iterations and learning rate for SGN and SGD respectively. A batch size of $1000$ is used for both SGD and SGN. Notice that the learning curve of SGD with learning rate 1 stops at epoch 7 as SGD starts diverging afterwards with this value of learning rate.}\label{fig:cifar10_1}
    \end{minipage}\qquad
    \begin{minipage}{0.8\textwidth}
        \centering
        {\includegraphics[width=\textwidth]{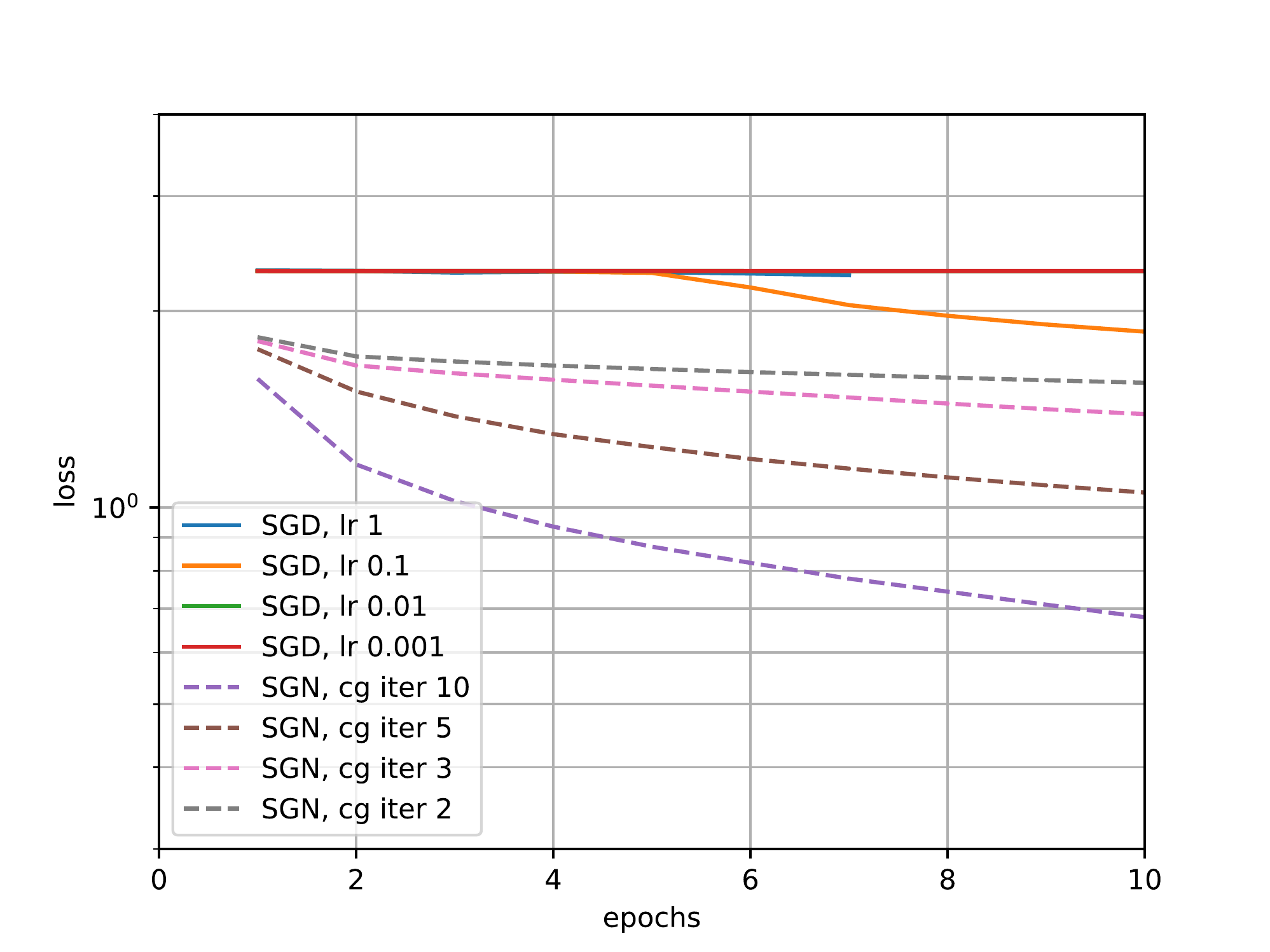}}
\caption{Average train loss vs. epochs over $5$ runs for SGN (dashed lines) and SGD (continuous lines) on CIFAR10. Different colors are used for different values of CG iterations and learning rate for SGN and SGD respectively. A batch size of $1000$ is used for both SGD and SGN. Notice that the learning curve of SGD with learning rate 1 stops at epoch 7 as SGD starts diverging afterwards with this value of learning rate.}\label{fig:cifar10_2}
    \end{minipage}
\end{figure}

\begin{figure}[!htb]
    \centering
    \begin{minipage}{.8\textwidth}
        \centering
       {\includegraphics[width=\textwidth]{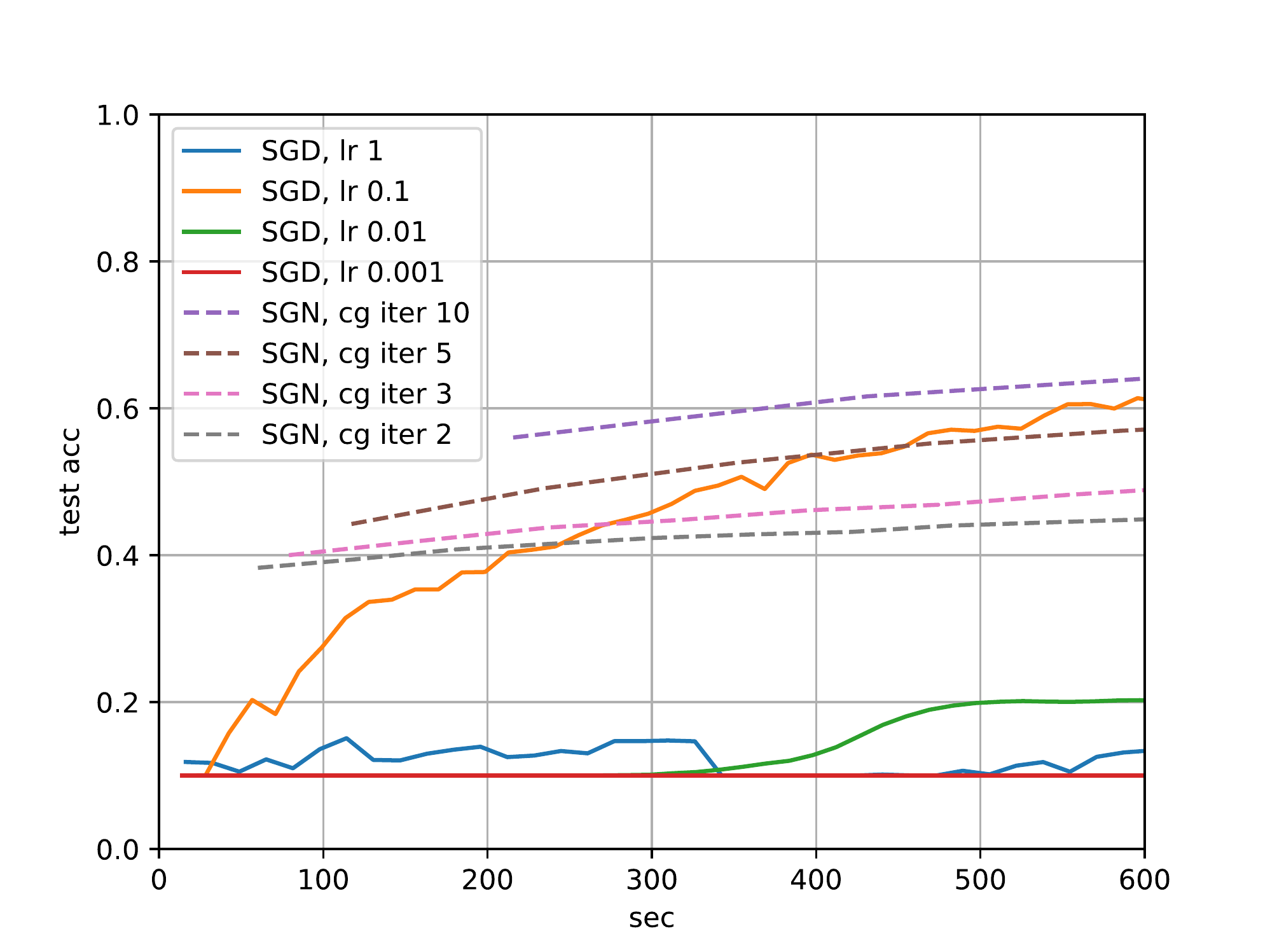}}
\caption{Average test accuracy vs. seconds over $5$ runs for SGN (dashed lines) and SGD (continuous lines) on CIFAR10. Different colors are used for different values of CG iterations and learning rate for SGN and SGD respectively. A batch size of $1000$ is used for both SGD and SGN.}\label{fig:cifar10_3}
    \end{minipage}\qquad
    \begin{minipage}{0.8\textwidth}
        \centering
        {\includegraphics[width=\textwidth]{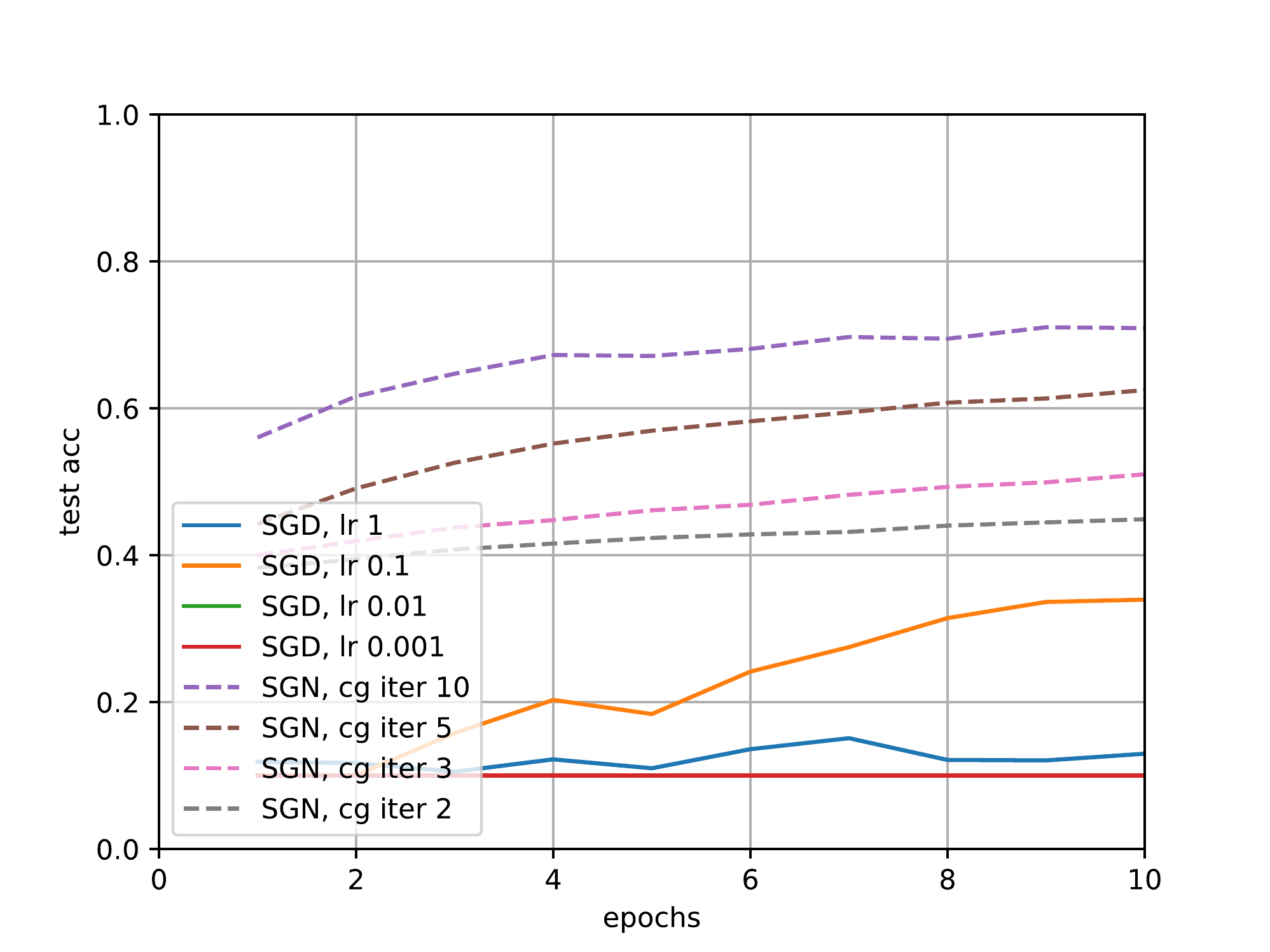}}
\caption{Average test accuracy vs. epochs over $5$ runs for SGN (dashed lines) and SGD (continuous lines) on CIFAR10. Different colors are used for different values of CG iterations and learning rate for SGN and SGD respectively. A batch size of $1000$ is used for both SGD and SGN.}\label{fig:cifar10_4}
    \end{minipage}
\end{figure}

\begin{figure}[!htb]
    \centering
    \begin{minipage}{.8\textwidth}
        \centering
       {\includegraphics[width=\textwidth]{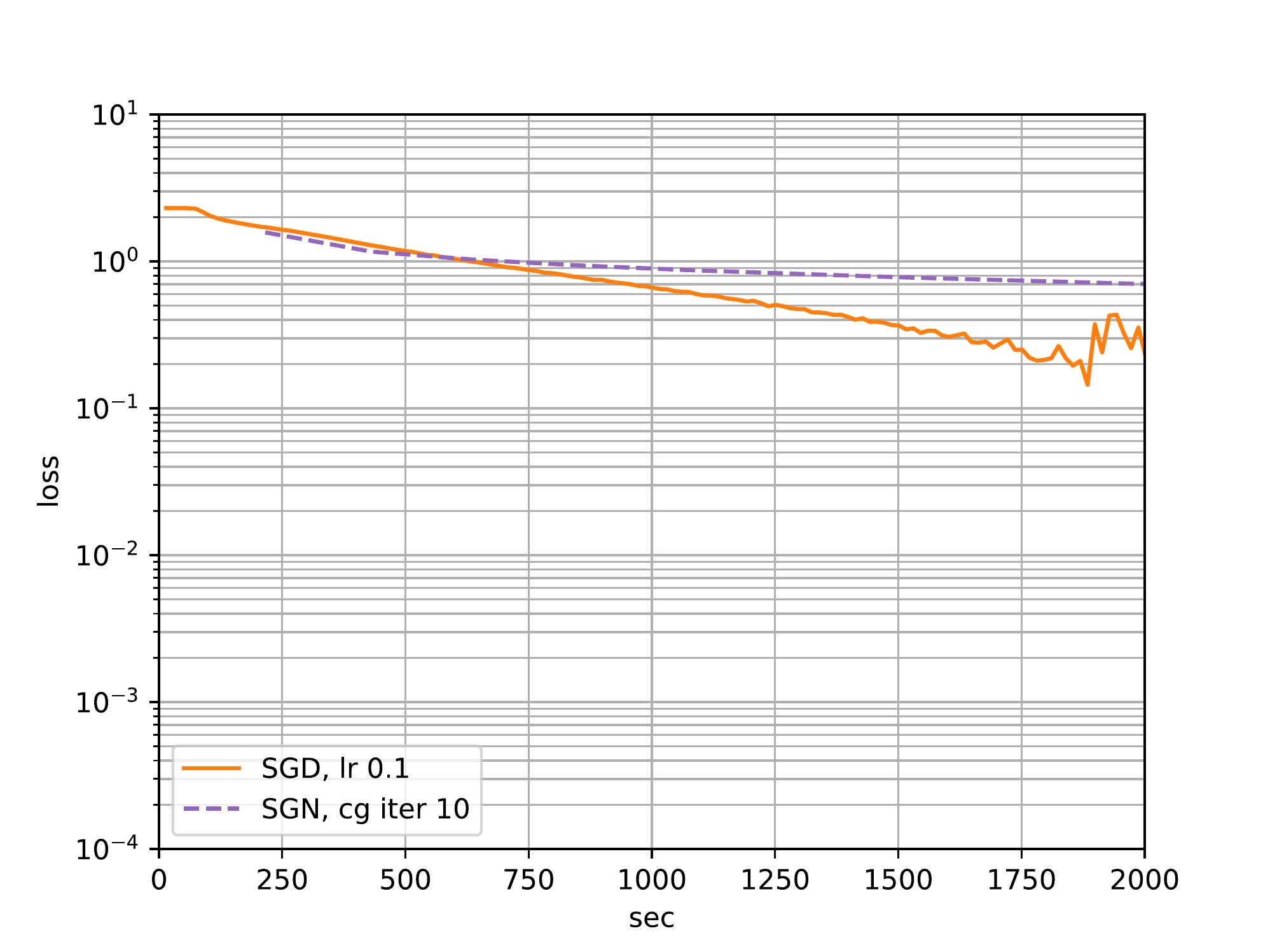}}
\caption{Average train loss vs. seconds over $5$ runs for SGN (dashed lines) and SGD (continuous lines) on CIFAR10. A batch size of $1000$ is used for both SGD and SGN.}\label{fig:cifar10_5}
    \end{minipage}\qquad
    \begin{minipage}{0.8\textwidth}
        \centering
        {\includegraphics[width=\textwidth]{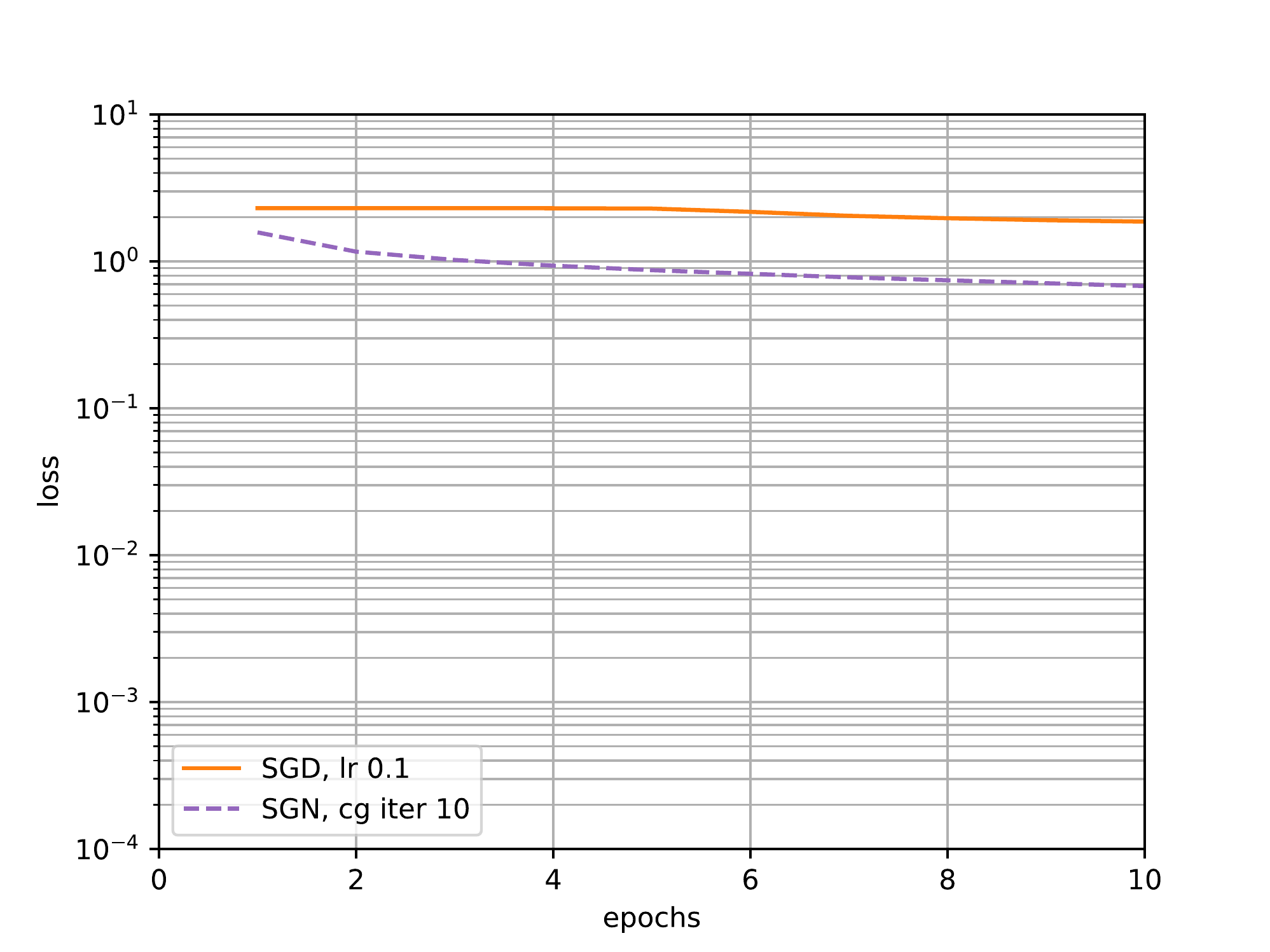}}
\caption{Average train loss vs. epochs over $5$ runs for SGN (dashed lines) and SGD (continuous lines) on CIFAR10. A batch size of $1000$ is used for both SGD and SGN.}\label{fig:cifar10_6}
    \end{minipage}
\end{figure}
\begin{figure}[!htb]
    \centering
    \begin{minipage}{.8\textwidth}
        \centering
       {\includegraphics[width=\textwidth]{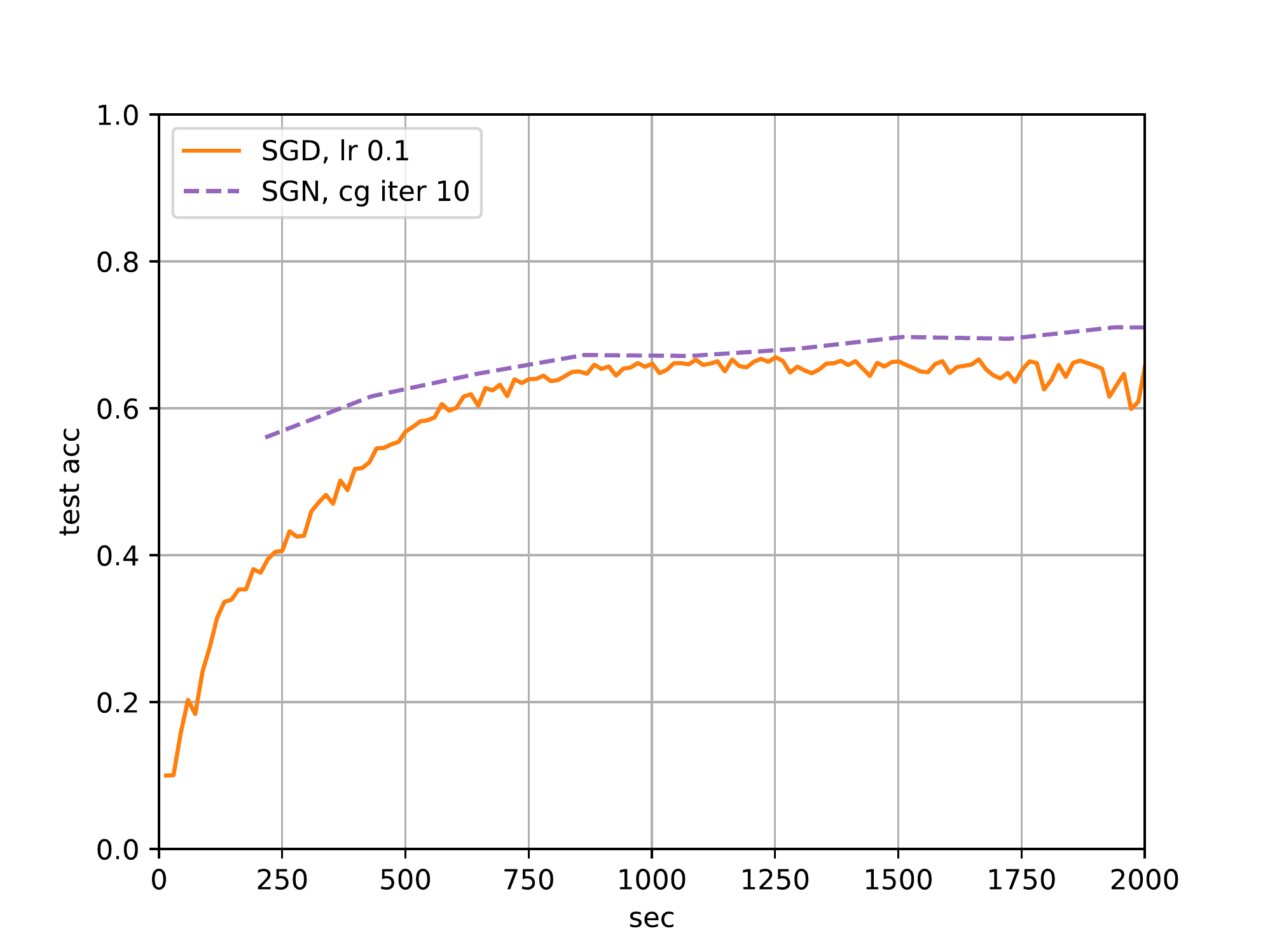}}
\caption{Average test accuracy vs. seconds over $5$ runs for SGN (dashed lines) and SGD (continuous lines) on CIFAR10. A batch size of $1000$ is used for both SGD and SGN.}\label{fig:cifar10_7}
    \end{minipage}\qquad
    \begin{minipage}{.8\textwidth}
        \centering
        {\includegraphics[width=\textwidth]{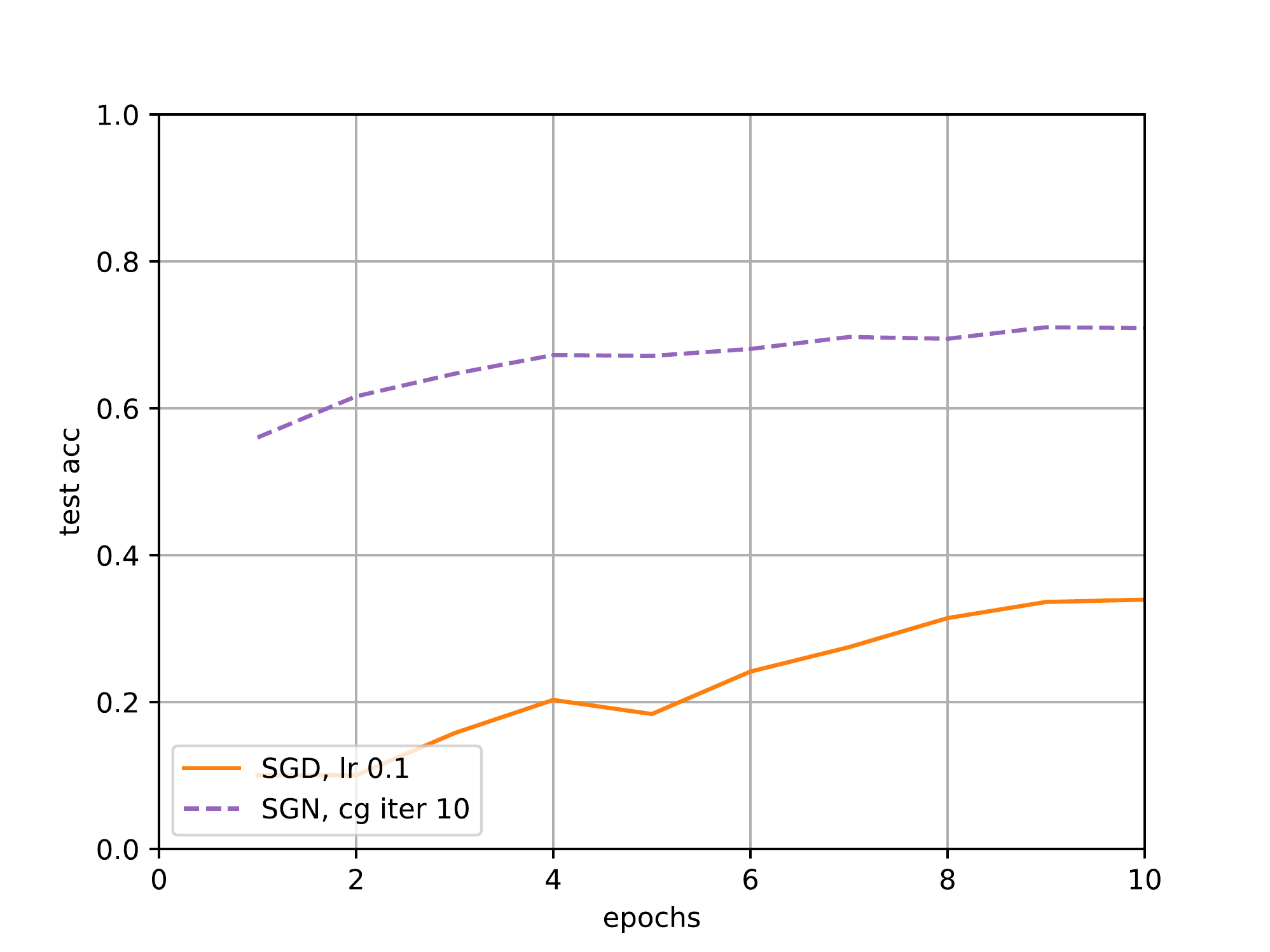}}
\caption{Average test accuracy vs. epochs over $5$ runs for SGN (dashed lines) and SGD (continuous lines) on CIFAR10. A batch size of $1000$ is used for both SGD and SGN.}\label{fig:cifar10_8}
    \end{minipage}
\end{figure}

\end{document}